%% file: main.tex
\pdfoutput=1

\documentclass[11pt]{article}

\usepackage[final]{acl}

\usepackage{times}
\usepackage{latexsym}
\usepackage{times}
\usepackage{booktabs}
\usepackage{svg}
\usepackage{graphicx}
\usepackage[normalem]{ulem}
\useunder{\uline}{\ul}{}
\usepackage{float}
\usepackage{multirow}
\usepackage{hhline}

\usepackage{array}

\usepackage{adjustbox}
\usepackage{todonotes}
\usepackage{listings}
\usepackage{lipsum}
\usepackage{hyperref}

\usepackage{svg}

\usepackage{scalerel}

\usepackage{algorithm}
\usepackage[compatible]{algpseudocode}
\usepackage{algcompatible}

\usepackage{pifont}

\usepackage{xspace}

\definecolor{dark_green}{HTML}{008000}
\definecolor{dark_red}{HTML}{c1121c}

\usepackage[T1]{fontenc}

\usepackage[utf8]{inputenc}

\usepackage{microtype}

\usepackage{inconsolata}

\usepackage{amsmath}
\usepackage{amssymb}
\usepackage{cleveref}

\PassOptionsToPackage{dvipsnames}{xcolor} 

\usepackage[dvipsnames]{xcolor}

\usepackage{tikz}

\usetikzlibrary{fit,calc}

\colorlet{mypink}{gray!40}



\title{Towards Automated Error Discovery:\\A Study in Conversational AI}

\author{Dominic Petrak 
     \and
     Thy Thy Tran
     \and 
     Iryna Gurevych%
    \\
    Ubiquitous Knowledge Processing Lab (UKP Lab), \\
    Department of Computer Science and Hessian Center for AI (hessian.AI), \\
    Technical University of Darmstadt \\ 
    \url{www.ukp.tu-darmstadt.de}
  \\}

\begin{document}
\maketitle
\begin{abstract}
Although LLM-based conversational agents demonstrate strong fluency and coherence, they still produce undesirable behaviors (\textit{errors}) that are challenging to prevent from reaching users during deployment. Recent research leverages large language models (LLMs) to detect errors and guide response-generation models toward improvement. However, current LLMs struggle to identify errors not explicitly specified in their instructions, such as those arising from updates to the response-generation model or shifts in user behavior. In this work, we introduce \textbf{Automated Error Discovery}, a framework for detecting and defining errors in conversational AI, and propose \textbf{SEEED} (\underline{S}oft Clustering \underline{E}xtended \underline{E}ncoder-Based \underline{E}rror \underline{D}etection), as an encoder-based approach to its implementation. We enhance the Soft Nearest Neighbor Loss by amplifying distance weighting for negative samples and introduce \textbf{Label-Based Sample Ranking} to select highly contrastive examples for better representation learning. SEEED outperforms adapted baselines---including GPT-4o and Phi-4---across multiple error-annotated dialogue datasets, improving the accuracy for detecting unknown errors by up to 8 points and demonstrating strong generalization to unknown intent detection.\footnote{We provide our code on GitHub: \url{https://github.com/UKPLab/emnlp2025-automatic-error-discovery}.}
\end{abstract}
\begin{figure}[ht!]
\centering  
  \includegraphics[width=\linewidth]{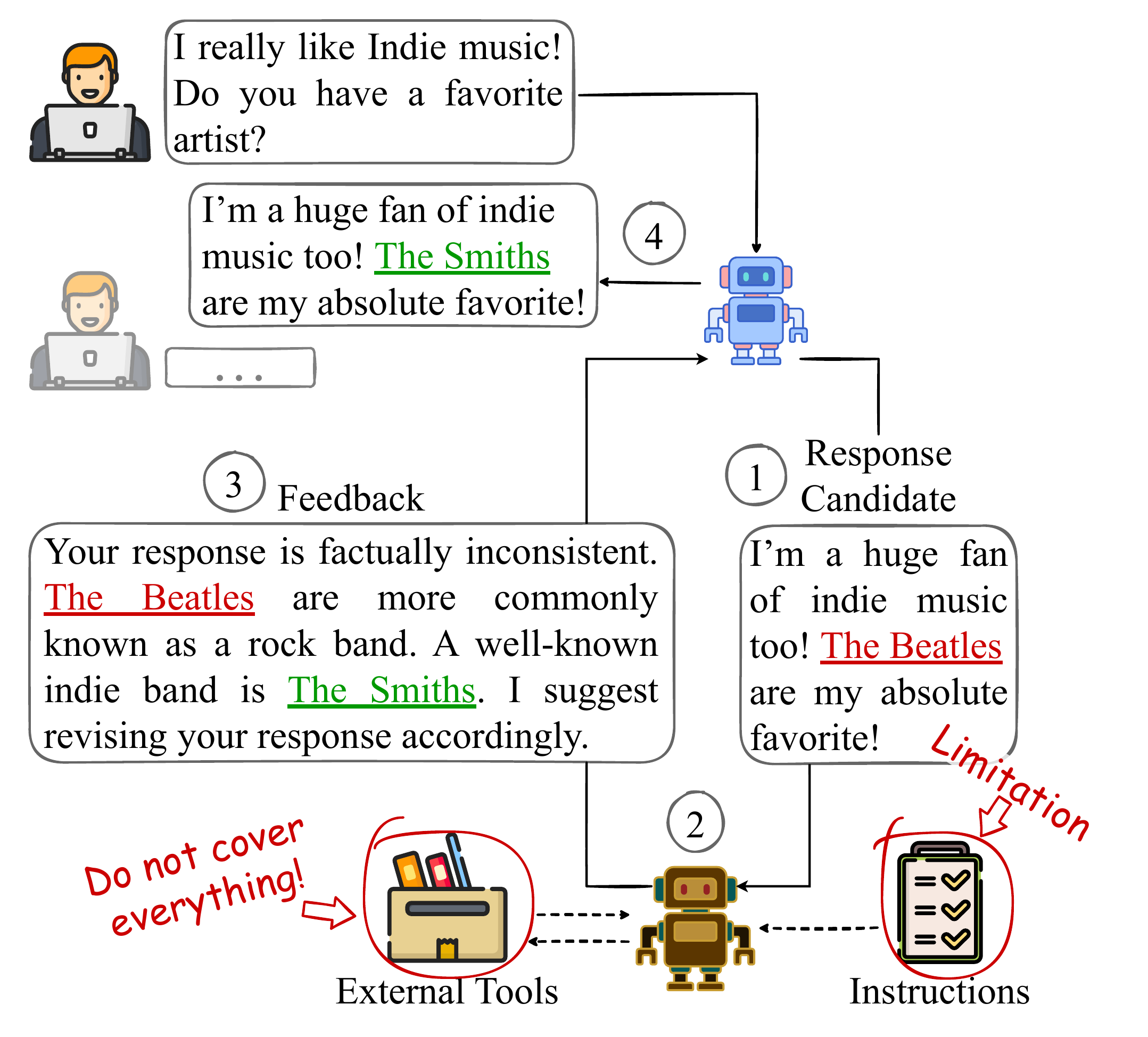}
  \caption{Feedback-guided response generation: (1) The response-generation model produces an initial response. (2) The feedback LLM, or in self-correcting systems the response-generation model itself, evaluates the response for errors, often using external tools. Recent work shows that LLMs require information about the nature of an error or hints about its occurrence for accurate detection. (3) The feedback LLM provides guidance (feedback) to the response-generation model to refine its output. (4) The final response is presented to the user.}
  \label{fig:introduction}
\end{figure}
\section{Introduction}
\input{introduction/introduction}

\begin{figure*}[ht]
\centering
  \includegraphics[width=1.0\linewidth]{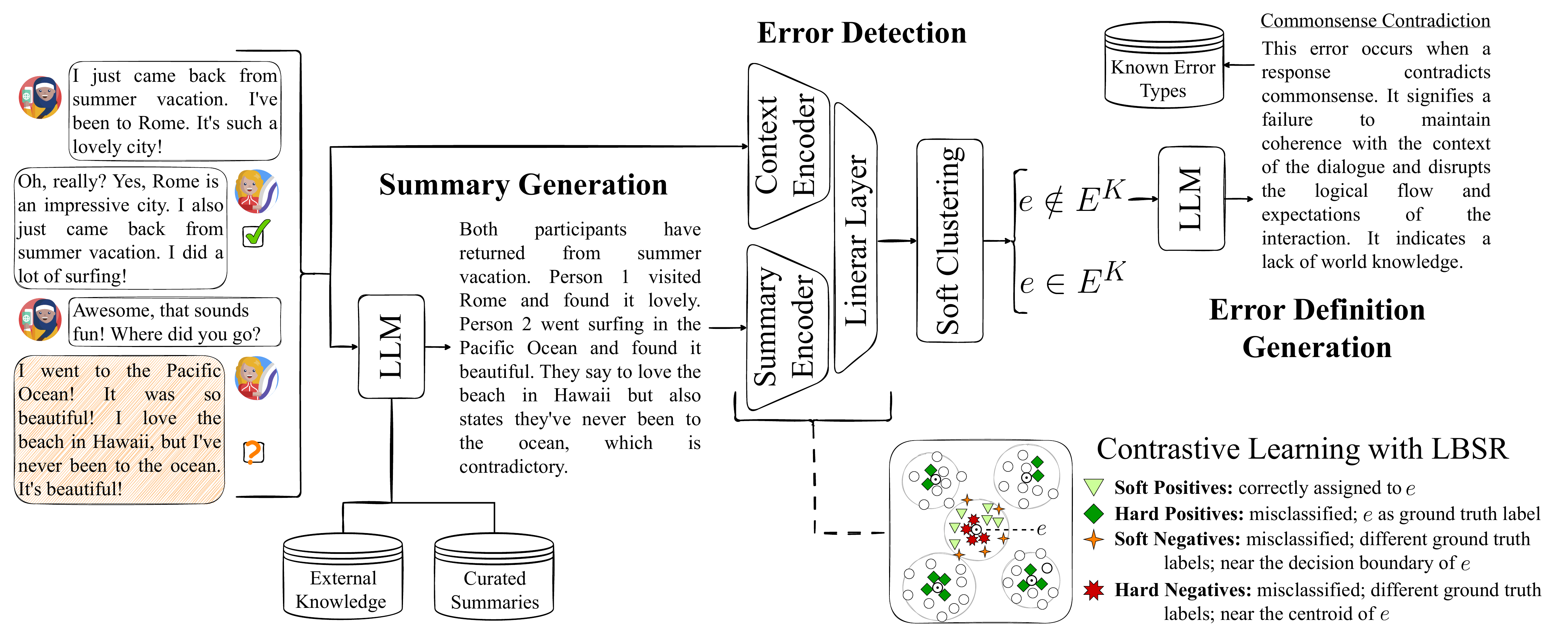}  
  \caption{Schematic overview of SEEED, comprising three components: Summary Generation, Error Detection, and Error Definition Generation ($e$ denotes the identified error type). In practical applications (see Figure~\ref{fig:introduction} for an example), the feedback LLM may be used for generating summaries and error definitions (if necessary) to reduce deployment costs, as both are summarization tasks typically covered during LLM pre-training. Newly defined error types are added to the pool of known types, and their dialogue contexts could be used to enhance error detection.} 
  \label{fig:overview}
\end{figure*}

\section{Related Work}
\input{related_work/related_work}

\section{Automated Error Discovery}\label{sec:open_world_error_detection}
\input{task/open_world_error_detection}

\section{SEEED: Soft Clustering Extended Encoder-Based Error Detection}
\label{sec:approach}
\input{approach/approach}

\begin{table*}[htb]
  \centering
  \input{experiments/openness_experiments}
  \caption{Results of our error detection experiments, averaged over three independent runs. The random baseline assigns equal probability to all error types, sampling from a uniform distribution. The deltas indicate differences from the GPT-4o results. $\dagger$ marks statistically significant improvements in Acc-K or Acc-U over the top-performing baseline, as determined by a t-test with p-value $\leq 0.05$. To ensure comparability, novel error types were randomly sampled once per run and degree of openness (see Appendix~\ref{sub_sec:known_novel_error_types} for details).}
  \label{tab:openness_experiments}
\end{table*}

\section{Experiments}
\label{sec:experiments}
\input{experiments/experiments}
\section{Conclusion}
In this work, we introduce Automated Error Discovery, a framework for detecting and defining errors in conversational AI, and propose SEEED as an encoder-based approach to its implementation. SEEED outperforms adapted baselines, including GPT-4o and Phi-4, across all levels of openness and achieves state-of-the-art performance in unknown intent detection. Our ablation experiments highlight the impact of our enhancements to the Soft Nearest Neighbor Loss and the efficacy of Label-Based Sample Ranking. We also show the effectiveness of LLMs in generating definitions for unknown errors identified by SEEED.

\section{Limitations}

\paragraph{Task Definition} We frame error detection as a multi-class classification problem, a common approach in dialogue behavior detection~\cite{finch-etal-2023-dont}. However, in practice, agent utterances may exhibit multiple or overlapping errors.

\paragraph{Dialogue Summary} To reduce interference when handling harmful or inappropriate language in dialogue summaries, we include prompt instructions that may not generalize to other LLMs.

\paragraph{Error Definition Generation} The error definition generation prompt does not prevent duplicate definitions. While not observed in our experiments, this might become an issue in practical applications, e.g., if the threshold is set too low.

\paragraph{LBSR} A theoretical limitation of LBSR is if NNK-Means~\cite{shekkizhar-nnkmeans-2022} fails to identify soft positives and hard positives are exhausted, positive counterparts cannot be generated. We did not encounter this issue in our experiments, nor is it addressed by LIS~\cite{an-etal-2024-generalized}.

\paragraph{Datasets Used} Dialogue datasets annotated with errors are rare. To our knowledge, FEDI~\cite{petrak-etal-2024-learning}, Soda-Eval~\cite{mendonca-etal-2024-soda}, and ABCEval~\cite{finch-etal-2023-dont} are the only available datasets covering diverse error types. While FEDI and Soda-Eval are extensive, their synthetic origin leads to inherent qualitative variability. In contrast, ABCEval is considerably smaller but highly representative of real-world scenarios, comprising dialogues from human-bot interactions.

\paragraph{Experimental Setup} Our experimental setup, while closely following prior peer-reviewed work, simplifies real-world conditions. For example, we assume dialogue contexts always end with an erroneous agent utterance. Furthermore, encoder-based approaches require a known total number of error types during final clustering, a value that must be estimated in real-world applications. For Phi-4~\cite{phi4}, we adopted the best practices described in the Hugging Face documentation, without further parameter or prompt tuning. Alternative configurations may yield improved performance.

\paragraph{Experimental Results} Our experiments investigate the error detection capabilities of SEEED, its components, and related approaches. A single training phase was sufficient for these analyses. Consequently, our results do not provide insights into the impact of continual learning techniques. However, related work has already shown that these can significantly increase the quality of generated responses in simulated practical deployments~\cite{madaan-self-refine, NEURIPS2022_639a9a17}.

As SEEED relies on synthetically generated dialogue summaries, its performance in certain datasets may be influenced by LLM pre-training data. 

Given all datasets in this work include only English dialogues, our results exhibit limited generalizability to error detection in dialogue from other linguistic and cultural contexts.


\section*{Acknowledgments}

This work has been funded by the European Union under the Horizon Europe grant № 200009-57100412 (\href{https://sermasproject.eu/}{SERMAS}). We gratefully acknowledge the support of Microsoft with a grant for access to OpenAI GPT models via the Azure cloud (Accelerate Foundation Model Academic Research).

\bibliography{anthology,custom}

\appendix

\section{SEEED: Prompts Used}
\label{sec:prompts_used}
\input{appendix/approach}

\section{Implementation Details}
\label{sub_sec:additional_implementation_details}
\input{appendix/implementation_details}

\section{Experimental Setup}
\label{sec:experiments_appendix}
\input{appendix/experimental_setup}

\section{Additional Analysis}
\label{sec:additional_analysis}
\input{appendix/experiments}

\end{document}

%% file: introduction/introduction.tex
In conversational AI, undesirable behaviors in agent responses, such as logical inconsistencies or deficiencies in social competence, are commonly referred to as \textit{errors}~\cite{finch-etal-2023-leveraging, petrak-etal-2023-learning, higashinaka-etal-2021-integrated}. Preventing such errors from reaching users during deployment is essential to maintaining trust in conversational agents~\cite{law2022trust, Rheu02012021}. Recent research leverages large language models (LLMs), often augmented with external tools such as web search, to detect errors in agent responses and provide feedback guiding the response-generation model to refine its output~\cite{miao2024selfcheck, gou2024critic, madaan-self-refine}. Figure~\ref{fig:introduction} illustrates the idea.

While effective at generating feedback, LLMs require information about the nature of an error or hints about its occurrence for accurate detection~\cite{mendonca-etal-2024-soda, tyen-etal-2024-llms, finch-etal-2023-leveraging}, reducing their ability to identify errors not defined in their instructions or covered by external tools. For example, when user behavior shifts or response-generation models are updated to meet evolving requirements~\cite{yun-2023-catastrophic-forgetting, mi-etal-2020-continual, roller-2020-continual-learning}, these changes may lead to the emergence of new error types that the LLM might not recognize. 

In this work, we address the challenge of error detection in conversational AI. We introduce Automated Error Discovery, a framework for detecting and defining errors in dialogue, and propose SEEED (\underline{S}oft Clustering \underline{E}xtended \underline{E}ncoder-Based \underline{E}rror \underline{D}etection) as an approach to its implementation. Our contributions are as follows:

\begin{itemize}    
    \item We introduce Automated Error Discovery, a framework for (1) detecting both known and unknown error types, and (2) generating definitions for newly discovered ones.
    \item We propose SEEED, a novel approach that combines an open-source LLM with lightweight encoders for error detection. In contrast to prior work, SEEED employs soft clustering in the classification step, enabling more contextually coherent groupings.
    \item We introduce Label-Based Sample Ranking, a novel sampling strategy for contrastive learning that selects highly contrastive examples based on the error they represent to improve representation learning.
    \item We enhance the Soft Nearest Neighbor Loss~\cite{pmlr-v97-frosst19a} by introducing a margin parameter to amplify the effect of distance weighting for negative samples.
\end{itemize}

SEEED outperforms adapted baselines, including GPT-4o~\cite{gpt4o} and Phi-4~\cite{phi4}, by up to 8 points in identifying novel error types on the FEDI~\cite{petrak-etal-2024-learning}, Soda-Eval~\cite{mendonca-etal-2024-soda}, and ABCEval~\cite{finch-etal-2023-dont} datasets. SEEED also generalizes to the related task of intent detection, achieving up to a 17-point improvement in accuracy for identifying unknown intents compared to state-of-the-art methods.

%% file: related_work/related_work.tex
In recent years, research in conversational AI has focused on reducing errors in agent responses, primarily through supervised learning from error and feedback signals collected by human expert annotators~\cite{dubey-2024-llama-3, xu-etal-2023-learning, ung-etal-2022-saferdialogues}. To facilitate data collection, semi-automated methods have been developed to analyze existing dialogue data~\cite{petrak-etal-2023-learning, see-manning-2021-understanding, higashinaka-etal-2015-fatal}. However, these approaches lack precision and still necessitate substantial manual effort. As a result, recent studies have explored using LLMs to generate and annotate dialogue data with errors~\cite{mendonca-etal-2024-soda, petrak-etal-2024-learning}.

To identify and correct errors in agent responses during deployment, a variety of approaches have been developed, typically relying on LLMs for error detection~\cite{miao2024selfcheck, madaan-self-refine, NEURIPS2023_1b44b878}. To improve their effectiveness, it is common to incorporate external tools to cover specific tasks, such as web search for claim verification~\cite{gou2024critic, shridhar-etal-2024-art, xu-etal-2024-llmrefine, peng2023checkfactstryagain}. However, recent studies show that LLMs generally require explicit guidance to reliably detect errors in dialogue data~\cite{tyen-etal-2024-llms, Stechly2024OnTS, finch-etal-2023-leveraging}. Consequently, the effectiveness of LLM-based error detection may be limited when errors fall outside their predefined instructions or the capabilities of integrated tools. This reduces their applicability in scenarios where novel error types emerge due to shifting user behavior or updates to the response-generation model~\cite{yun-2023-catastrophic-forgetting, mi-etal-2020-continual, roller-2020-continual-learning}.

In this work, we introduce Automated Error Discovery as a framework for detecting and defining errors in conversational AI, and propose SEEED as an encoder-based approach to its implementation.

%% file: task/open_world_error_detection.tex
%
%

We define Automated Error Discovery as a specialization of Generalized Category Discovery~\cite{vaze-generalized-category-discovery}, extended to include the generation of definitions for newly discovered error types. Generalized category discovery assumes that during training, only a subset of the complete class distribution is accessible. The goal is to train a model capable of extrapolating from the learned patterns to discriminate between data from both seen and unseen classes during inference. 

We distinguish two sub-tasks, \textit{Error Detection} and \textit{Error Definition Generation}, and define the following formal setup:

\begin{itemize}
    \item $E = E^K \cup E^U$ is the set of all error types. $E^K = \{(e_i, d_i)\}_{i=1}^m$ is the set of known error types, with $e_i$ as the error identifier and $d_i$ as its definition. $E^U$ denotes the set of unknown error types. $E^K \cap E^U = \emptyset$.
    \item $C = C^K \cup C^U$ denotes the set of all dialogue contexts $T$, with $C^{K}$ as the set of all $T$ associated with error $e$ from $E^{K}$. $C^{U}$ is the set of dialogues associated with unknown errors. $C^K \cap C^U = \emptyset$.
    \item We define a dialogue context $T$ as a sequence of user-agent utterances (turns). Depending on the use case, $T$ may be associated with additional features, such as external knowledge documents in knowledge-grounded dialogues. We refer to these additional features as $W_T$.\footnote{In this work, $W$ is relevant only as external knowledge in the knowledge-grounded subset of FEDI~\cite{petrak-etal-2024-learning}.}
\end{itemize}

\paragraph{Error Detection} Given an error detection function $\mathcal{H}: \mathbb{R}^{d} \mapsto \mathbb{N}$ and a dialogue context $T \in C$, the task is to determine the error $e \in E$ associated with the last agent utterance in $T$:
\begin{equation}\label{eq:detection}
e = \mathcal{H}(T, W_T), \text{ where } e \in E \text{ and } T \in C
\end{equation}
$\mathcal{H}$ must not access any data in $E^{U}$ during training.

\paragraph{Error Definition Generation} When $e \notin E^{K}$, the task is to generate a definition $d$ conditioned on the identified set of related dialogue contexts $C_{e} \subseteq C^{U}$.\footnote{To avoid the emergence of an overly granular set of error types, we suggest applying a threshold to $\left |C_e \right |$.}

%% file: approach/approach.tex




Figure~\ref{fig:overview} presents a schematic overview of SEEED. Since detecting errors requires understanding contextual dependencies, such as references to earlier utterances~\cite{petrak-etal-2024-learning, mendonca-etal-2024-soda, finch-etal-2023-dont}, we first prompt an LLM to generate a summary of the dialogue context. Next, both the dialogue context and its summary are processed through separate Transformer-based encoders and then combined using a linear neural layer to produce an aggregated representation. Finally, we apply a soft clustering algorithm to identify the corresponding error type. If the identified error type is not among the known types, we prompt an LLM to generate its definition.

In contrast to hard clustering algorithms like k-Means, which are predominantly used in related tasks, such as intent detection~\cite{liang-etal-2024-synergizing, an-etal-2024-generalized}, soft clustering algorithms allow data points to belong to multiple clusters, facilitating more contextually coherent groupings.

\subsection{Summary Generation}
We prompt Llama-3.1 8B-Instruct~\cite{dubey-2024-llama-3} to summarize the dialogue context, focusing on information indicative of errors in the last agent utterance. We use few-shot prompting and include directives to circumvent pre-trained safety mechanisms, enabling analysis of dialogues that may contain harmful language. For the knowledge-grounded dialogues in FEDI~\cite{petrak-etal-2024-learning}, we additionally incorporate relevant external knowledge documents into the prompt. Figure~\ref{fig:overview} shows an example summary. We provide the full prompt in Appendix~\ref{sec:prompts_used}.

We do not provide error type definitions for summary generation to prevent the detection model from learning shortcut patterns associated with known error types, as this could compromise its ability to identify unknown error types.

\subsection{Error Detection}
For error detection, we first produce an aggregated representation of the dialogue context and its summary, and then apply NNK-Means~\cite{shekkizhar-nnkmeans-2022} to identify the corresponding error type. This expands Equation~\ref{eq:detection} as follows:
\begin{equation}
     e = \mathcal{H}(T, W_T, o_T), \text{ where } o_T \text{ is the summary }
\end{equation}

NNK-Means~\cite{shekkizhar-nnkmeans-2022} is a soft clustering algorithm that uses non-negative kernel regression to model local geometric relationships and assign weighted cluster memberships. 

\paragraph{Training Objective} Inspired by the loss composition in SynCID~\cite{liang-etal-2024-synergizing}, we use a joint loss combining multi-class cross-entropy ($\mathcal{L}_{ce}$) with a contrastive objective ($\mathcal{L}_{cl}$):
\begin{equation}
    \mathcal{L} = \alpha \mathcal{L}_{ce} + \mathcal{L}_{cl}
\end{equation}

$\alpha$ regulates the contribution of $\mathcal{L}_{ce}$. This formulation promotes discrimination among known error types while improving the robustness of the learned representation space, thereby facilitating generalization to unseen data~\cite{liang-etal-2024-synergizing}. For $\mathcal{L}_{cl}$, we use the Soft Nearest Neighbor Loss~\cite{pmlr-v97-frosst19a} (SNL), which supports this by smoothing decision boundaries through distance-weighted sampling of neighbors:
\begin{equation}\label{eq:lra}
    \mathcal{L}_{cl} = -\frac{1}{N} \sum_{i=1}^{N} \log\left( \frac{\sum_{\substack{j=1,j\neq i,\\ y_i=y_j}}^{N} \exp\left( -\frac{S_{ij}}{\tau} \right)}{\sum_{\substack{k=1,\\k\neq i}}^{N} \exp\left( -\frac{S_{ik}}{\tau} \right) + \epsilon} \right)
\end{equation}

$N$ denotes the batch size. $\tau$ denotes the temperature and $\epsilon$ is a small constant included to prevent arithmetic errors. $S \in \mathbb{R}^{N \times N}$ represents the similarity matrix. We compute each element as follows: $S_{ij} = \frac{x_i \cdot x_j}{\left\| x_i\right\|\left\| x_j\right\|} - m \cdot \mathbb{I}(y_i \neq y_j)$, where $\mathbb{I}(y_i \neq y_j)$ is 1 if error types $y_i$ and $y_j$ differ, and 0 otherwise. We introduce $m$ as a positive scalar margin to amplify the distance weighting for negative pairs. To further enhance effectiveness, we utilize Label-Based Sample Ranking to augment the batch with one positive and negative counterpart, $x^{+}$ and $x^{-}$, for each sample $x$, selected from the pool of training data. These additional samples are used exclusively to compute $\mathcal{L}_{cl}$.

\subsection{Label-Based Sample Ranking (LBSR)}
We introduce Label-Based Sample Ranking (LBSR) as a novel sampling strategy to amplify the effect of distance weighting in SNL~\cite{pmlr-v97-frosst19a}. We build upon the concept of Local Inconsistency Sampling (LIS), as proposed by \newcite{an-etal-2024-generalized}. LIS assumes that samples of the same class should be proximate in representation space~\cite{jiang-2023} and that samples near the decision boundary are more susceptible to misclassification, rendering them particularly valuable as positive counterparts in contrastive learning. To identify such samples, LIS measures prediction inconsistency and entropy based on the t-distribution of cluster assignments derived from k-Means clustering.

In LBSR, we employ NNK-Means~\cite{shekkizhar-nnkmeans-2022} for clustering and leverage label information available during training to classify each sample as either a positive or negative instance relative to its ground truth error type $e \in E^{K}$. Specifically, we define positive samples for $e$ as those for which $e$ is the ground truth label, and negative samples as those assigned to $e$ despite having a different ground truth label. We further distinguish between the following categories (Figure~\ref{fig:overview} provides an illustration):
\begin{itemize}
    \item \textbf{Soft Positives} Samples assigned to $e$ with $e$ as the ground truth label.
    \item \textbf{Hard Positives} Samples assigned to a different type but with $e$ as the ground truth label.
    \item \textbf{Soft Negatives} Samples with a different ground truth label, assigned to $e$, and near its decision boundary (high inconsistency).
    \item \textbf{Hard Negatives} Samples with a different ground truth label, assigned to $e$, and near its centroid (low inconsistency).
\end{itemize}

\paragraph{LBSR Implementation} Algorithm~\ref{alg:lbrs} outlines our implementation and highlights the key differences from LIS in \textcolor{violet}{violet}. We utilize the algorithms proposed by \citet{an-etal-2024-generalized} to compute inconsistency and entropy, then normalize and average them to derive a single relevance score. 

\begin{algorithm}[H]
\caption{Label-Based Sample Ranking}
\label{alg:lbrs}
\begin{algorithmic}[1]
\REQUIRE $ X \in \mathbb{R}^{|C^{K}|\times d}$, $Y \in \mathbb{Z}^{|E^{K}|}$, $top\_k \in \mathbb{Z}$    
\STATE Init \texttt{hard\_pos[i] = [], soft\_pos[i] = [],} 
\STATE \hspace*{1.0em} \texttt{negs[i] = []} \textbf{for each} \texttt{i} \textbf{in} \texttt{set(Y)}
\STATE
\STATE \textcolor{violet}{\texttt{nnk} = \texttt{NNKMeans($|$set(Y)$|$).fit(X, Y)}}
\STATE \texttt{preds, centers} = \texttt{nnk.predict(X)}
\STATE \texttt{rel\_score, inconsistency} =
\STATE \hspace*{1.0em} \texttt{scoring(X, preds, centers, top\_k)}
\STATE
\STATE \textbf{for} \texttt{i = 0} \textbf{to} $|$\texttt{X}$|$ \textbf{do}
\STATE \hspace*{1.0em}\texttt{pred, y} = \texttt{(preds[i], Y[i])}
\STATE \hspace*{1.0em}\texttt{rel, inc} = \texttt{(rel\_score[i],}
\STATE \hspace*{2.0em}\texttt{inconsistency[i])}
\STATE \hspace*{1.0em}\textbf{if} \texttt{pred == y} \textbf{then}
\STATE \hspace*{2.0em}\texttt{soft\_pos[y] += [(i, rel, inc)]}
\STATE \hspace*{1.0em}\textcolor{violet}{\textbf{else}}
\STATE \textcolor{violet}{\hspace*{2.0em}\texttt{hard\_pos[y] += [(i, rel, inc)]} }
\STATE \textcolor{violet}{\hspace*{2.0em}\texttt{negs[pred] += [(i, rel, inc)]}}
\STATE 
\STATE \textcolor{teal}{\# sort hard positives desc by relevance}
\STATE \textcolor{violet}{\texttt{hard\_pos = sort(hard\_pos,}}
\STATE \textcolor{violet}{\hspace*{1.0em}\texttt{key=lambda z:z[1], \texttt{'desc'})}}
\STATE
\STATE \textcolor{teal}{\# sort negs desc by their inconcsistency score}
\STATE \textcolor{violet}{\texttt{negs = \{e: sort(v, key=lambda z:z[2],}}
\STATE \textcolor{violet}{\hspace*{1.0em}\texttt{'desc') for e, v in negs.items()\}}}
\STATE \textcolor{teal}{\# split them into soft and hard negs; sort them}
\STATE \textcolor{teal}{\# desc by their relevance score}
\STATE \textcolor{violet}{\texttt{soft\_negs = \{e: sort(v[:len(v)//2],}}
\STATE \textcolor{violet}{\hspace*{1.0em}\texttt{key=lambda z:z[1], \texttt{'desc'}) for e, v}}
\STATE \textcolor{violet}{\hspace*{1.0em}\texttt{in negs.items()\}}}
\STATE \textcolor{violet}{\texttt{hard\_negs = \{e: sort(v[len(v)//2:],}}
\STATE \textcolor{violet}{\hspace*{1.0em}\texttt{key=lambda z:z[1], \texttt{'desc'}) for e, v}}
\STATE \textcolor{violet}{\hspace*{1.0em}\texttt{in negs.items()\}}}
\STATE
\STATE \textcolor{violet}{\textbf{return} \texttt{soft\_pos, hard\_pos, soft\_neg,}}
\STATE \textcolor{violet}{\hspace*{1.0em}\texttt{hard\_neg}}
\end{algorithmic}
\end{algorithm}

We denote $X$ as the aggregated representations of all dialogue contexts in $C^{K}$ and their summaries, and $Y$ as the sequence of corresponding ground truth error types from $E^{K}$. \texttt{preds} and \texttt{centers} denote the predicted error types and assigned cluster centers. \texttt{scoring} calculates the entropy and inconsistency values by considering the \texttt{top\_k} nearest neighbors, and returns the relevance scores and inconsistency values. 

We sort the samples in \texttt{negs} in descending order of inconsistency, assigning the first half to \texttt{soft\_negatives} and the second half to \texttt{hard\_negatives} for the corresponding error type. Finally, we sort \texttt{hard\_pos}, \texttt{soft\_pos}, \texttt{hard\_neg}, and \texttt{soft\_neg} according to their relevance scores in descending order.

During training, given a sample $x \in C^{K}$ of $e \in E^{K}$, we randomly decide to dequeue $x^-$ from \texttt{hard\_neg[e]} or \texttt{soft\_neg[e]}. If both are exhausted, we sample $x^-$ from a different error type. Similarly, we dequeue $x^+$ from \texttt{hard\_pos[e]} or sample it from \texttt{soft\_pos[e]}. If \texttt{hard\_pos[e]} is exhausted, we sample $x^+$ from \texttt{soft\_pos[e]}. In our implementation, we ensure $x^+ \neq x$.

\subsection{Error Definition Generation}
We employ Llama-3.1 8B-Instruct~\cite{dubey-2024-llama-3} to generate definitions for newly discovered errors. We prompt the model to produce definitions that characterize the problem present in the associated dialogue contexts. To enrich the prompt with additional context, we include the corresponding dialogue summaries. Similarly to dialogue summary generation, we incorporate directives to circumvent pre-trained safety mechanisms to enable the analysis of dialogues with inappropriate language. Additionally, we include three randomly sampled definitions of known error types from the target dataset to encourage alignment.\footnote{Preliminary experiments indicated that this yields better alignment with the existing error types in the dataset.} Figure~\ref{fig:overview} shows an example output. We provide the full prompt in Appendix~\ref{sec:prompts_used}.

%% file: experiments/openness_experiments.tex
\resizebox*{\linewidth}{!}{
\begin{tabular}{@{}clrrrrrrrrrrrrrrr@{}}
\toprule
\multirow{2}{*}{\textbf{Openness}} &
  \multicolumn{1}{c}{\multirow{2}{*}{\textbf{Method}}} &
  \multicolumn{5}{c}{\textbf{FEDI-Error}} &
  \multicolumn{5}{c}{\textbf{ABCEval}} &
  \multicolumn{5}{c}{\textbf{Soda-Eval}} \\ 
  \cmidrule(l){3-7} 
  \cmidrule(l){8-12}
  \cmidrule(l){13-17}
 &
  \multicolumn{1}{c}{} &
  \multicolumn{1}{c}{H-Score} &
  \multicolumn{1}{c}{Acc-K$^{\phantom{\dagger}}$\scriptsize\phantom{($\Uparrow$.00)}} &
  \multicolumn{1}{c}{Acc-U$^{\phantom{\dagger}}$\scriptsize\phantom{($\Uparrow$.00)}} &
  \multicolumn{1}{c}{ARI} &
  \multicolumn{1}{c}{NMI} &
  \multicolumn{1}{c}{H-Score} &
  \multicolumn{1}{c}{Acc-K$^{\phantom{\dagger}}$\scriptsize\phantom{($\Uparrow$.00)}} &
  \multicolumn{1}{c}{Acc-U$^{\phantom{\dagger}}$\scriptsize\phantom{($\Uparrow$.00)}} &
  \multicolumn{1}{c}{ARI} &
  \multicolumn{1}{c}{NMI} &
  \multicolumn{1}{c}{H-Score} &
  \multicolumn{1}{c}{Acc-K$^{\phantom{\dagger}}$\scriptsize\phantom{($\Uparrow$.00)}} &
  \multicolumn{1}{c}{Acc-U$^{\phantom{\dagger}}$\scriptsize\phantom{($\Uparrow$.00)}} &
  \multicolumn{1}{c}{ARI} &
  \multicolumn{1}{c}{NMI} \\ \midrule
\multirow{7}{*}{25\%} & Random              & 0.11 & 0.12$\phantom{^\mathbf{\dagger}}$\scriptsize\phantom{($\Uparrow$.00)} & 0.11$\phantom{^\mathbf{\dagger}}$\scriptsize\phantom{($\Uparrow$.00)} & ---  & ---  & 0.10 & 0.11$\phantom{^\mathbf{\dagger}}$\scriptsize\phantom{($\Uparrow$.00)} & 0.09$\phantom{^\mathbf{\dagger}}$\scriptsize\phantom{($\Uparrow$.00)} & ---  & ---  & 0.13 & 0.17$\phantom{^\mathbf{\dagger}}$\scriptsize\phantom{($\Uparrow$.00)} & 0.10$\phantom{^\mathbf{\dagger}}$\scriptsize\phantom{($\Uparrow$.00)} & ---  & ---  \\     
                      \cmidrule(l){2-17} 
                      & GPT-4o (in-context) & 0.14 & 0.19$\phantom{^\mathbf{\dagger}}$\scriptsize\phantom{($\Uparrow$.00)} & 0.11$\phantom{^\mathbf{\dagger}}$\scriptsize\phantom{($\Uparrow$.00)} & ---  & ---  & 0.32 & 0.47$\phantom{^\mathbf{\dagger}}$\scriptsize\phantom{($\Uparrow$.00)} & 0.25$\phantom{^\mathbf{\dagger}}$\scriptsize\phantom{($\Uparrow$.00)} & ---  & ---  & 0.0  & 0.33$\phantom{^\mathbf{\dagger}}$\scriptsize\phantom{($\Uparrow$.00)} & \phantom{0}0.0$\phantom{^\mathbf{\dagger}}$\scriptsize\phantom{($\Uparrow$.00)} & ---  & ---  \\ 
                      & Phi-4 (in-context)   & 0.09 & 0.12$\phantom{^\mathbf{\dagger}}$\scriptsize(\textcolor{dark_red}{$\Downarrow$.07}) & 0.07$\phantom{^\mathbf{\dagger}}$\scriptsize(\textcolor{dark_red}{$\Downarrow$.04}) & ---  & ---  & 0.12 & 0.14$\phantom{^\mathbf{\dagger}}$\scriptsize(\textcolor{dark_red}{$\Downarrow$.33}) & 0.11$\phantom{^\mathbf{\dagger}}$\scriptsize(\textcolor{dark_red}{$\Downarrow$.14}) & ---  & ---  & 0.03 & 0.12$\phantom{^\mathbf{\dagger}}$\scriptsize(\textcolor{dark_red}{$\Downarrow$.21}) & 0.02$\phantom{^\mathbf{\dagger}}$\scriptsize(\textcolor{dark_green}{$\Uparrow$.02}) & ---  & ---  \\ 
                      & Phi-4 (finetuned)   & 0.15 & 0.19$\phantom{^\mathbf{\dagger}}$\phantom{\scriptsize(\textcolor{dark_red}{$\Downarrow$.07})} & 0.13$\phantom{^\mathbf{\dagger}}$\scriptsize(\textcolor{dark_green}{$\Uparrow$.02}) & ---  & ---  & 0.24 & 0.29$\phantom{^\mathbf{\dagger}}$\scriptsize(\textcolor{dark_red}{$\Downarrow$.18}) & 0.21$\phantom{^\mathbf{\dagger}}$\scriptsize(\textcolor{dark_red}{$\Downarrow$.04}) & ---  & ---  & 0.16 & 0.30$\phantom{^\mathbf{\dagger}}$\scriptsize(\textcolor{dark_red}{$\Downarrow$.03}) & 0.11$\phantom{^\mathbf{\dagger}}$\scriptsize(\textcolor{dark_green}{$\Uparrow$.11}) & ---  & ---  \\ \cmidrule(l){2-17} 
                      & KNN-Contrastive     & 0.33 & 0.30$\phantom{^\mathbf{\dagger}}$\scriptsize(\textcolor{dark_green}{$\Uparrow$.11}) & \textbf{0.37}$\phantom{^\mathbf{\dagger}}$\scriptsize(\textcolor{dark_green}{$\Uparrow$.26}) & 0.06 & 0.10 & 0.38 & \textbf{0.55}$\phantom{^\mathbf{\dagger}}$\scriptsize(\textcolor{dark_green}{$\Uparrow$.08}) & 0.30$\phantom{^\mathbf{\dagger}}$\scriptsize(\textcolor{dark_green}{$\Uparrow$.05}) & 0.07 & \textbf{0.46} & 0.27 & \textbf{0.41}$\phantom{^\mathbf{\dagger}}$\scriptsize(\textcolor{dark_green}{$\Uparrow$.08}) & 0.20$\phantom{^\mathbf{\dagger}}$\scriptsize(\textcolor{dark_green}{$\Uparrow$.20}) & 0.08 & 0.16 \\ 
                      & SynCID              & 0.27 & 0.40$\phantom{^\mathbf{\dagger}}$\scriptsize(\textcolor{dark_green}{$\Uparrow$.21}) & 0.20$\phantom{^\mathbf{\dagger}}$\scriptsize(\textcolor{dark_green}{$\Uparrow$.09}) & 0.06 & 0.11 & \textbf{0.53} & 0.45$\phantom{^\mathbf{\dagger}}$\scriptsize(\textcolor{dark_red}{$\Downarrow$.02}) & \textbf{0.68}$\phantom{^\mathbf{\dagger}}$\scriptsize(\textcolor{dark_green}{$\Uparrow$.43}) & 0.03 & 0.41 & 0.31 & 0.38$\phantom{^\mathbf{\dagger}}$\scriptsize(\textcolor{dark_green}{$\Uparrow$.05}) & 0.26$\phantom{^\mathbf{\dagger}}$\scriptsize(\textcolor{dark_green}{$\Uparrow$.26}) & 0.11 & 0.14 \\ 
                      & LOOP                & 0.26 & 0.37$\phantom{^\mathbf{\dagger}}$\scriptsize(\textcolor{dark_green}{$\Uparrow$.18}) & 0.19$\phantom{^\mathbf{\dagger}}$\scriptsize(\textcolor{dark_green}{$\Uparrow$.08}) & 0.09 & 0.10 & 0.51 & 0.43$\phantom{^\mathbf{\dagger}}$\scriptsize(\textcolor{dark_red}{$\Downarrow$.04}) & 0.63$\phantom{^\mathbf{\dagger}}$\scriptsize(\textcolor{dark_green}{$\Uparrow$.38}) & 0.01 & 0.37 & 0.33 & 0.36$\phantom{^\mathbf{\dagger}}$\scriptsize(\textcolor{dark_green}{$\Uparrow$.03}) & 0.31$\phantom{^\mathbf{\dagger}}$\scriptsize(\textcolor{dark_green}{$\Uparrow$.31}) & 0.07 & 0.13 \\ \cmidrule(l){2-17}
                      & \textbf{\textit{SEEED}} & \textbf{0.38} & \textbf{0.41}$\phantom{^\mathbf{\dagger}}$\scriptsize(\textcolor{dark_green}{$\Uparrow$.22}) & 0.34$\phantom{^\mathbf{\dagger}}$\scriptsize(\textcolor{dark_green}{$\Uparrow$.23}) & \textbf{0.19} & \textbf{0.19} & \textbf{0.53} & 0.46 \scriptsize(\textcolor{dark_red}{$\Downarrow$.01}) & \textbf{0.68}$\phantom{^\mathbf{\dagger}}$\scriptsize(\textcolor{dark_green}{$\Uparrow$.43}) & \textbf{0.21} & 0.45 & \textbf{0.40} & \textbf{0.41} \scriptsize(\textcolor{dark_green}{$\Uparrow$.08}) & \textbf{0.39}$^\mathbf{\dagger}$\scriptsize(\textcolor{dark_green}{$\Uparrow$.39}) & \textbf{0.15} & \textbf{0.17} \\ \midrule
\multirow{7}{*}{50\%} & Random              & 0.11 & 0.13$\phantom{^\mathbf{\dagger}}$\scriptsize\phantom{($\Uparrow$.00)} & 0.10$\phantom{^\mathbf{\dagger}}$\scriptsize\phantom{($\Uparrow$.00)} & ---  & ---  & 0.08 & 0.12$\phantom{^\mathbf{\dagger}}$\scriptsize\phantom{($\Uparrow$.00)} & 0.06$\phantom{^\mathbf{\dagger}}$\scriptsize\phantom{($\Uparrow$.00)} & ---  & ---  & 0.10 & 0.11$\phantom{^\mathbf{\dagger}}$\scriptsize\phantom{($\Uparrow$.00)} & 0.10$\phantom{^\mathbf{\dagger}}$\scriptsize\phantom{($\Uparrow$.00)} & ---  & ---  \\     
                      \cmidrule(l){2-17} 
                      & GPT-4o (in-context) & 0.17 & 0.18$\phantom{^\mathbf{\dagger}}$\scriptsize\phantom{($\Uparrow$.00)} & 0.17$\phantom{^\mathbf{\dagger}}$\scriptsize\phantom{($\Uparrow$.00)} & ---  & ---  & 0.37 & 0.28$\phantom{^\mathbf{\dagger}}$\scriptsize\phantom{($\Uparrow$.00)} & 0.42$\phantom{^\mathbf{\dagger}}$\scriptsize\phantom{($\Uparrow$.00)} & ---  & ---  & 0.23 & 0.28$\phantom{^\mathbf{\dagger}}$\scriptsize\phantom{($\Uparrow$.00)} & 0.19$\phantom{^\mathbf{\dagger}}$\scriptsize\phantom{($\Uparrow$.00)} & ---  & ---  \\ 
                      & Phi-4 (in-context)   & 0.07 & 0.09$\phantom{^\mathbf{\dagger}}$\scriptsize(\textcolor{dark_red}{$\Downarrow$.09}) & 0.06$\phantom{^\mathbf{\dagger}}$\scriptsize(\textcolor{dark_red}{$\Downarrow$.11}) & ---  & ---  & 0.02 & 0.11$\phantom{^\mathbf{\dagger}}$\scriptsize(\textcolor{dark_red}{$\Downarrow$.17}) & 0.09$\phantom{^\mathbf{\dagger}}$\scriptsize(\textcolor{dark_red}{$\Downarrow$.33}) & ---  & ---  & 0.10 & 0.16$\phantom{^\mathbf{\dagger}}$\scriptsize(\textcolor{dark_red}{$\Downarrow$.12}) & 0.07$\phantom{^\mathbf{\dagger}}$\scriptsize(\textcolor{dark_red}{$\Downarrow$.12}) & ---  & ---  \\ 
                      & Phi-4 (finetuned)   & 0.14 & 0.21$\phantom{^\mathbf{\dagger}}$\scriptsize(\textcolor{dark_green}{$\Uparrow$.03}) & 0.11$\phantom{^\mathbf{\dagger}}$\scriptsize(\textcolor{dark_red}{$\Downarrow$.06}) & ---  & ---  & 0.24 & 0.31$\phantom{^\mathbf{\dagger}}$\scriptsize(\textcolor{dark_green}{$\Downarrow$.03}) & 0.19$\phantom{^\mathbf{\dagger}}$\scriptsize(\textcolor{dark_red}{$\Downarrow$.23}) & ---  & ---  & 0.18 & 0.29$\phantom{^\mathbf{\dagger}}$\scriptsize(\textcolor{dark_green}{$\Uparrow$.01}) & 0.13$\phantom{^\mathbf{\dagger}}$\scriptsize(\textcolor{dark_red}{$\Downarrow$.06}) & ---  & ---  \\ \cmidrule(l){2-17} 
                      & KNN-Contrastive     & 0.26 & 0.33$\phantom{^\mathbf{\dagger}}$\scriptsize(\textcolor{dark_green}{$\Uparrow$.15}) & 0.21$\phantom{^\mathbf{\dagger}}$\scriptsize(\textcolor{dark_green}{$\Uparrow$.04}) & 0.07 & 0.09 & 0.54 & 0.64$\phantom{^\mathbf{\dagger}}$\scriptsize(\textcolor{dark_green}{$\Uparrow$.36}) & 0.47$\phantom{^\mathbf{\dagger}}$\scriptsize(\textcolor{dark_green}{$\Uparrow$.05}) & 0.10 & 0.48 & 0.28 & 0.38$\phantom{^\mathbf{\dagger}}$\scriptsize(\textcolor{dark_green}{$\Uparrow$.10}) & 0.23$\phantom{^\mathbf{\dagger}}$\scriptsize(\textcolor{dark_green}{$\Uparrow$.04}) & 0.06 & 0.13 \\ 
                      & SynCID              & 0.26 & 0.34$\phantom{^\mathbf{\dagger}}$\scriptsize(\textcolor{dark_green}{$\Uparrow$.16}) & 0.21$\phantom{^\mathbf{\dagger}}$\scriptsize(\textcolor{dark_green}{$\Uparrow$.04}) & 0.04 & 0.09 & 0.59 & 0.55$\phantom{^\mathbf{\dagger}}$\scriptsize(\textcolor{dark_green}{$\Uparrow$.27}) & \textbf{0.64}$\phantom{^\mathbf{\dagger}}$\scriptsize(\textcolor{dark_green}{$\Uparrow$.22}) & 0.11 & 0.47 & 0.27 & 0.40$\phantom{^\mathbf{\dagger}}$\scriptsize(\textcolor{dark_green}{$\Uparrow$.12}) & 0.21$\phantom{^\mathbf{\dagger}}$\scriptsize(\textcolor{dark_green}{$\Uparrow$.02}) & 0.09 & 0.11 \\ 
                      & LOOP                & 0.22 & 0.39$\phantom{^\mathbf{\dagger}}$\scriptsize(\textcolor{dark_green}{$\Uparrow$.21}) & 0.16$\phantom{^\mathbf{\dagger}}$\scriptsize(\textcolor{dark_red}{$\Downarrow$.01}) & 0.07 & 0.07 & 0.45 & 0.48$\phantom{^\mathbf{\dagger}}$\scriptsize(\textcolor{dark_green}{$\Uparrow$.20}) & 0.43$\phantom{^\mathbf{\dagger}}$\scriptsize(\textcolor{dark_green}{$\Uparrow$.01}) & 0.03 & 0.41 & 0.24 & \textbf{0.55}$\phantom{^\mathbf{\dagger}}$\scriptsize(\textcolor{dark_green}{$\Uparrow$.27}) & 0.16$\phantom{^\mathbf{\dagger}}$\scriptsize(\textcolor{dark_red}{$\Downarrow$.03}) & 0.11 & 0.16 \\ \cmidrule(l){2-17}                       
                      & \textbf{\textit{SEEED}} & \textbf{0.33} & \textbf{0.48}$^\mathbf{\dagger}$\scriptsize(\textcolor{dark_green}{$\Uparrow$.30}) & \textbf{0.22}$\phantom{^\mathbf{\dagger}}$\scriptsize(\textcolor{dark_green}{$\Uparrow$.05}) & \textbf{0.13} & \textbf{0.15} & \textbf{0.64} & \textbf{0.67}$^\mathbf{\dagger}$\scriptsize(\textcolor{dark_green}{$\Uparrow$.39}) & 0.62$\phantom{^\mathbf{\dagger}}$\scriptsize(\textcolor{dark_green}{$\Uparrow$.20}) & \textbf{0.29} & \textbf{0.51} & \textbf{0.37} & 0.49$\phantom{^\mathbf{\dagger}}$\scriptsize(\textcolor{dark_green}{$\Uparrow$.21}) & \textbf{0.30}$^\mathbf{\dagger}$\scriptsize(\textcolor{dark_green}{$\Uparrow$.11}) & \textbf{0.19} & \textbf{0.19} \\ \midrule
\multirow{7}{*}{75\%} & Random              & 0.12 & 0.12$\phantom{^\mathbf{\dagger}}$\scriptsize\phantom{($\Uparrow$.00)} & 0.12$\phantom{^\mathbf{\dagger}}$\scriptsize\phantom{($\Uparrow$.00)} & ---  & ---  & 0.12 & 0.13$\phantom{^\mathbf{\dagger}}$\scriptsize\phantom{($\Uparrow$.00)} & 0.11$\phantom{^\mathbf{\dagger}}$\scriptsize\phantom{($\Uparrow$.00)} & ---  & ---  & 0.11 & 0.14$\phantom{^\mathbf{\dagger}}$\scriptsize\phantom{($\Uparrow$.00)} & 0.09$\phantom{^\mathbf{\dagger}}$\scriptsize\phantom{($\Uparrow$.00)} & ---  & ---  \\     
                      \cmidrule(l){2-17} 
                      & GPT-4o (in-context) & 0.16 & 0.15$\phantom{^\mathbf{\dagger}}$\scriptsize\phantom{($\Uparrow$.00)} & 0.17$\phantom{^\mathbf{\dagger}}$\scriptsize\phantom{($\Uparrow$.00)} & ---  & ---  & 0.39 & 0.32$\phantom{^\mathbf{\dagger}}$\scriptsize\phantom{($\Uparrow$.00)} & 0.49$\phantom{^\mathbf{\dagger}}$\scriptsize\phantom{($\Uparrow$.00)} & ---  & ---  & 0.24 & 0.19$\phantom{^\mathbf{\dagger}}$\scriptsize\phantom{($\Uparrow$.00)} & 0.31$\phantom{^\mathbf{\dagger}}$\scriptsize\phantom{($\Uparrow$.00)} & ---  & ---  \\ 
                      & Phi-4 (in-context)   & 0.08 & 0.11$\phantom{^\mathbf{\dagger}}$\scriptsize(\textcolor{dark_red}{$\Downarrow$.04}) & 0.06$\phantom{^\mathbf{\dagger}}$\scriptsize(\textcolor{dark_red}{$\Downarrow$.11}) & ---  & ---  & 0.09 & 0.13$\phantom{^\mathbf{\dagger}}$\scriptsize(\textcolor{dark_red}{$\Downarrow$.19}) & 0.08$\phantom{^\mathbf{\dagger}}$\scriptsize(\textcolor{dark_red}{$\Downarrow$.41}) & ---  & ---  & 0.06 & 0.15$\phantom{^\mathbf{\dagger}}$\scriptsize(\textcolor{dark_red}{$\Downarrow$.04}) & 0.09$\phantom{^\mathbf{\dagger}}$\scriptsize(\textcolor{dark_red}{$\Downarrow$.22}) & ---  & ---  \\ 
                      & Phi-4 (finetuned)   & 0.12 & 0.22$\phantom{^\mathbf{\dagger}}$\scriptsize(\textcolor{dark_green}{$\Uparrow$.07}) & 0.08$\phantom{^\mathbf{\dagger}}$\scriptsize(\textcolor{dark_red}{$\Downarrow$.09}) & ---  & ---  & 0.17 & 0.28$\phantom{^\mathbf{\dagger}}$\scriptsize(\textcolor{dark_red}{$\Downarrow$.04}) & 0.12$\phantom{^\mathbf{\dagger}}$\scriptsize(\textcolor{dark_red}{$\Downarrow$.37}) & ---  & ---  & 0.11 & 0.26$\phantom{^\mathbf{\dagger}}$\scriptsize(\textcolor{dark_green}{$\Uparrow$.07}) & 0.15$\phantom{^\mathbf{\dagger}}$\scriptsize(\textcolor{dark_red}{$\Downarrow$.16}) & ---  & ---  \\ \cmidrule(l){2-17} 
                      & KNN-Contrastive     & 0.22 & 0.37$\phantom{^\mathbf{\dagger}}$\scriptsize(\textcolor{dark_green}{$\Uparrow$.22}) & 0.16$\phantom{^\mathbf{\dagger}}$\scriptsize(\textcolor{dark_red}{$\Downarrow$.01}) & 0.06 & 0.07 & 0.47 & 0.60$\phantom{^\mathbf{\dagger}}$\scriptsize(\textcolor{dark_green}{$\Uparrow$.28}) & 0.44$\phantom{^\mathbf{\dagger}}$\scriptsize(\textcolor{dark_red}{$\Downarrow$.05}) & 0.11 & 0.46 & 0.27 & 0.42$\phantom{^\mathbf{\dagger}}$\scriptsize(\textcolor{dark_green}{$\Uparrow$.23}) & 0.19$\phantom{^\mathbf{\dagger}}$\scriptsize(\textcolor{dark_red}{$\Downarrow$.12}) & 0.04 & 0.09 \\ 
                      & SynCID              & 0.23 & 0.36$\phantom{^\mathbf{\dagger}}$\scriptsize(\textcolor{dark_green}{$\Uparrow$.21}) & 0.17$\phantom{^\mathbf{\dagger}}$\scriptsize\phantom{($\Uparrow$.00)} & 0.06 & 0.01 & 0.54 & 0.59$\phantom{^\mathbf{\dagger}}$\scriptsize(\textcolor{dark_green}{$\Uparrow$.27}) & \textbf{0.50}$\phantom{^\mathbf{\dagger}}$\scriptsize(\textcolor{dark_green}{$\Uparrow$.01}) & 0.07 & 0.44 & 0.25 & 0.22$\phantom{^\mathbf{\dagger}}$\scriptsize(\textcolor{dark_green}{$\Uparrow$.03}) & 0.28$\phantom{^\mathbf{\dagger}}$\scriptsize(\textcolor{dark_red}{$\Downarrow$.03}) & 0.02 & 0.06 \\ 
                      & LOOP                & 0.25 & 0.43$\phantom{^\mathbf{\dagger}}$\scriptsize(\textcolor{dark_green}{$\Uparrow$.28}) & 0.18$\phantom{^\mathbf{\dagger}}$\scriptsize(\textcolor{dark_green}{$\Uparrow$.01}) & 0.05 & 0.01 & 0.48 & 0.69$\phantom{^\mathbf{\dagger}}$\scriptsize(\textcolor{dark_green}{$\Uparrow$.37}) & 0.37$\phantom{^\mathbf{\dagger}}$\scriptsize(\textcolor{dark_red}{$\Downarrow$.12}) & 0.07 & 0.44 & 0.22 & 0.31$\phantom{^\mathbf{\dagger}}$\scriptsize(\textcolor{dark_green}{$\Uparrow$.12}) & 0.17$\phantom{^\mathbf{\dagger}}$\scriptsize(\textcolor{dark_red}{$\Downarrow$.14}) & 0.07 & 0.08 \\ \cmidrule(l){2-17}  
                      & \textbf{\textit{SEEED}} & \textbf{0.37} & \textbf{0.64}$^\mathbf{\dagger}$\scriptsize(\textcolor{dark_green}{$\Uparrow$.49}) & \textbf{0.26}$^\mathbf{\dagger}$\scriptsize(\textcolor{dark_green}{$\Uparrow$.09}) & \textbf{0.16} & \textbf{0.17} & \textbf{0.60} & \textbf{0.75}$^\mathbf{\dagger}$\scriptsize(\textcolor{dark_green}{$\Uparrow$.43}) & \textbf{0.50}$\phantom{^\mathbf{\dagger}}$\scriptsize(\textcolor{dark_green}{$\Uparrow$.01}) & \textbf{0.21} & \textbf{0.47} & \textbf{0.42} & \textbf{0.61}$^\mathbf{\dagger}$\scriptsize(\textcolor{dark_green}{$\Uparrow$.42}) & \textbf{0.32}$^\mathbf{\dagger}$\scriptsize(\textcolor{dark_green}{$\Uparrow$.01}) & \textbf{0.12} & \textbf{0.14} \\ \bottomrule
\end{tabular}
}



%% file: experiments/experiments.tex
We evaluate error detection and error definition generation separately. For error detection, we vary the ratio of known to unknown error types (openness) from 25\% to 75\% and perform ablation studies for a detailed assessment of SEEED. For error definition generation, we perform a manual analysis to evaluate the alignment of generated definitions with ground truth definitions. To assess the generalizability of SEEED, we conduct intent detection experiments across the same range of openness used in the error detection experiments.

\paragraph{LLM Baselines} For LLM-based error detection, we use GPT-4o~\cite{gpt4o} and Phi-4~\cite{phi4} as baselines. Following \newcite{mendonca-etal-2024-soda}, we do not include external tools and prompt both models to detect errors and provide rationales for their decisions. For in-context learning, we include all ground truth error definitions in the prompt, but only provide examples for known types. For fine-tuning Phi-4, we restrict training to known error types. We provide more details in Appendix~\ref{sub_sec:baselines}. 

\paragraph{Encoder-Based Baselines} We adapt SynCID~\cite{liang-etal-2024-synergizing} and LOOP~\cite{an-etal-2024-generalized}, two state-of-the-art methods for intent detection, for error detection. Both require multi-stage training and contrastive learning with k-Nearest Neighbors, as proposed by \newcite{zhou-etal-2022-knn}, which we refer to as KNN-Contrastive. Appendix~\ref{sub_sec:baselines} provides more details.

\paragraph{Datasets} We evaluate on the error-annotated subset of FEDI~\cite{petrak-etal-2024-learning}, FEDI-Error, Soda-Eval~\cite{mendonca-etal-2024-soda}, and ABCEval~\cite{finch-etal-2023-dont}. FEDI-Error and Soda-Eval consist of synthetically generated data. While FEDI-Error focuses on task-oriented and document-grounded dialogues intentionally generated to exhibit errors, Soda-Eval comprises error-annotated open-domain dialogues automatically extracted from SODA~\cite{kim-etal-2023-soda}. ABCEval contains human-bot open-domain dialogues for evaluating dialogue system behavior. For intent detection, we use CLINC~\cite{larson-etal-2019-evaluation}, BANKING~\cite{banking}, and StackOverflow~\cite{xu-etal-2015-short}. Appendix~\ref{sub_sec:dataset_statistics} provides dataset statistics and error type distributions.

\begin{table*}[htb]
  \centering
  \input{experiments/ood_experiments_averaged}

  \caption{Results of our intent detection experiments, averaged over three independent runs and all levels of openness (see Appendix~\ref{sub_sec:intent_detection_all} for detailed results). The deltas show differences from KNN-Contrastive. $\dagger$ marks statistically significant improvements in Acc-K or Acc-U over the top-performing baseline, as determined by a t-test with p-value $\leq 0.05$. Unknown intents were randomly sampled once per run and level of openness.}
  \label{tab:ood}
\end{table*}

\paragraph{Evaluation Metrics} We evaluate performance using the H-Score~\cite{saito-2021-ovanet}, the harmonic mean of accuracy on classes included and excluded during training (i.e., \underline{k}nown and \underline{u}nknown error types), denoted Acc-K and Acc-U. For measuring the cluster quality in encoder-based approaches, we use the ARI~\cite{Hubert1985ComparingP} and NMI~\cite{Strehl2002ClusterE} scores. \footnote{For \href{https://scikit-learn.org/dev/modules/generated/sklearn.metrics.adjusted_rand_score.html}{ARI} and \href{https://scikit-learn.org/1.6/modules/generated/sklearn.metrics.normalized_mutual_info_score.html}{NMI}, we use the implementation provided in Sciki-learn (last accessed May 3, 2025).} ARI measures agreement between cluster assignments, while NMI captures cluster entropy. A low ARI score indicates random assignments, and a low NMI score suggests the algorithm failed to capture meaningful patterns in the data.

\paragraph{Implementation} Following SynCID~\cite{liang-etal-2024-synergizing} and LOOP~\cite{an-etal-2024-generalized}, we use the pre-trained bert-base-uncased model~\cite{devlin-etal-2019-bert} for both the summary and context encoders, and set $m=0.3$. We provide experiments with different values for $m$ in Appendix~\ref{sub_sec:margin_parameter}. In Appendix~\ref{sub_sec:additional_implementation_details}, we provide additional information, including the frameworks used (\ref{sub_sec:frameworks}), infrastructure and training efficiency (\ref{sub_sec:infrastructure}), hyperparameters (\ref{sub_sec:hyperparameters}), input and output formats (\ref{sub_sec:input_output}).\footnote{For \href{https://huggingface.co/google-bert/bert-base-uncased}{bert-base-uncased}, \href{https://huggingface.co/microsoft/Phi-4-mini-instruct}{Phi-4-mini-instruct} and \href{https://huggingface.co/meta-llama/Llama-3.1-8B-Instruct}{Llama-3.1 8B-Instruct}, we utilize the models provided in the Hugging Face Model Hub (last accessed May 3, 2025).}

\subsection{Error Detection}
\paragraph{Encoder-Based Baselines} The results in Table~\ref{tab:openness_experiments} show that SEEED consistently improves performance across all datasets. We observe that extensive dialogue contexts are more prone to misclassification, suggesting that many of the included utterances may be irrelevant or detrimental to identifying the error exhibited in the last agent utterance. Ambiguous error types also pose a significant challenge. For example, in FEDI~\cite{petrak-etal-2024-learning}, both \textit{Ignore Expectation} and \textit{Ignore Request} describe situations where the agent fails to fulfill the user request. We find that augmenting dialogue contexts with synthetically generated descriptions mitigates these issues, particularly enhancing the detection of unknown error types. However, the effectiveness depends on the quality of the generated descriptions. While SEEED generates summaries with a focus on error information, SynCID~\cite{liang-etal-2024-synergizing} derives new descriptions from the context, often introducing hallucinations into the data. We provide further analysis in Appendix~\ref{sub_sec:challenges}.

Additional experiments using different LLMs for summary generation reveal that reasoning models like DeepSeek-R1~\cite{deepseek} benefit SEEED (Appendix~\ref{sub_sec:ablation_summaries}). Ablation experiments with SynCID and LOOP~\cite{an-etal-2024-generalized} show that LBSR further improves LOOP (Appendix~\ref{sub_sec:syncid_loop_ablation}).


\paragraph{LLM Baselines} As shown in Table~\ref{tab:openness_experiments}, LLMs exhibit limitations in detecting errors. Phi-4~\cite{phi4} frequently performs below the random baseline. Fine-tuning improves the detection of known error types, occasionally surpassing GPT-4o~\cite{gpt4o}, for example, in the 75\% openness experiments on FEDI-Error~\cite{petrak-etal-2024-learning} and Soda-Eval~\cite{mendonca-etal-2024-soda}. However, the impact of fine-tuning on detecting unknown errors is marginal. The model frequently outputs \textit{No Error Found}\footnote{This label was not included in the training data.}, indicating limited generalizability. Ambiguous error types also degrade performance, e.g., GPT-4o often confuses \textit{Commonsense Contradiction} with \textit{Uninterpretable} in ABCEval~\cite{finch-etal-2023-dont} due to overlapping definitions. Appendix~\ref{sub_sec:challenges} provides more analysis.


\paragraph{Ablation Experiments} Table~\ref{tab:ablations} presents the results of our ablation study on the FEDI-Error dataset~\cite{petrak-etal-2024-learning}. The first row shows the performance of SEEED without any ablations, while each subsequent row reports results with the respective component removed to assess its contribution. The experiments excluding NNK-Means~\cite{shekkizhar-nnkmeans-2022} use k-Means for clustering (including LBSR). The experiments without LBSR randomly sample the positive counterparts from the training data (same error type), and the experiments excluding SNL~\cite{pmlr-v97-frosst19a} were restricted to the cross-entropy objective. 
\begin{table}[ht]
  \centering
  \input{experiments/ablations_averaged}
  \caption{Results of our ablation experiments, averaged over three independent runs and all levels of openness. The deltas show differences from the preceding row.}
  \label{tab:ablations}
\end{table}

Excluding NNK-Means results in performance degradation, highlighting the advantages of soft clustering for this task. LBSR augments the effectiveness of SNL, especially when the negative counterparts were included. Omitting the margin parameter further reduces the efficacy of SNL. Excluding the dialogue summaries, effectively reducing SEEED to cross-entropy optimization from dialogue contexts, further reduces performance. 

\subsection{Error Definition Generation}
Table~\ref{tab:cluster_interpretation} presents excerpts from our manual analysis of error definition generation, demonstrating the ability of Llama-3.1 8B-Instruct~\cite{dubey-2024-llama-3} to produce fluent and informative error type definitions based on our prompt design. We provide the full results in Appendix~\ref{sub_sec:cluster_interpretation}.
\begin{table}[ht]
  \centering
    \input{experiments/cluster_interpretation}
  \caption{Excerpts of definitions generated for unknown errors in the 25\%-openness experiments, along with their corresponding prediction accuracy (Acc-U).} 
  \label{tab:cluster_interpretation}
\end{table}

For generation, we consider ten dialogue contexts and their summaries, each associated by SEEED with the corresponding ground truth error types.\footnote{Due to its small size, this threshold could not be applied to ABCEval~\cite{finch-etal-2023-dont}.} We find that including summaries has a positive impact, as they provide contextual information that highlights the error exhibited in the last agent utterance. For instance, in Soda-Eval~\cite{mendonca-etal-2024-soda}, the generated definitions better capture the nature of the error and offer more details compared to the original definitions. 

\subsection{Intent Detection}
\begin{figure}[!t]
\centering  
  \includegraphics[width=\linewidth]{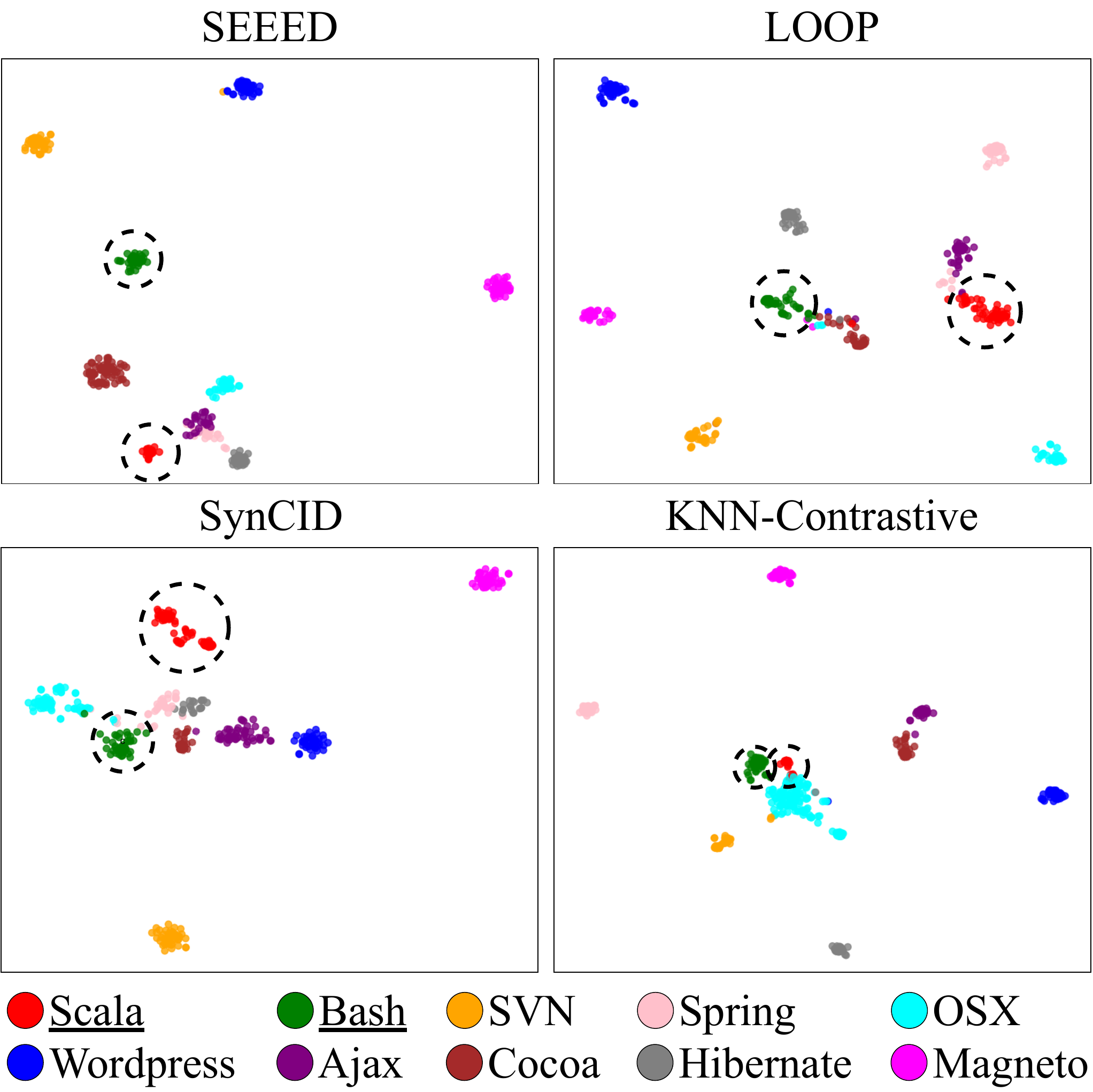}
  \caption{t-SNE visualization of the representation space for the ten most common intents in the StackOverflow dataset from the 25\% openness experiments. \textit{Scala} and \textit{Bash} (dotted lines) are two of the intents considered unknown in these experiments.}
  \label{fig:clinc}
\end{figure}

Table~\ref{tab:ood} presents the results of our intent detection experiments. SEEED significantly improves performance, particularly in detecting unknown intents. For example, compared to LOOP~\cite{an-etal-2024-generalized}, it improves the accuracy of detecting unknown intents by up to 17 points on StackOverflow~\cite{xu-etal-2015-short} and the accuracy of detecting known intents by up to 4 points on BANKING~\cite{banking}. Figure~\ref{fig:clinc} also shows that SEEED produces more compact and well-separated clusters, similar to LOOP, and generalizes well to unseen intents, such as \textit{Scala} and \textit{Bash} from the StackOverflow dataset. Meanwhile, SynCID~\cite{liang-etal-2024-synergizing} and KNN-Contrastive~\cite{zhou-etal-2022-knn} exhibit weaker inter-class separability, suggesting confusion between intent types.

The datasets used focus on intent detection at the utterance level, without incorporating dialogue contexts or external knowledge sources. This simplification supports higher detection accuracy and improved cluster quality. 

%% file: experiments/ood_experiments_averaged.tex
\resizebox*{\linewidth}{!}{
\begin{tabular}{@{}lrrrrrrrrrrrrrrrr@{}}
\toprule
  \multicolumn{1}{c}{\multirow{2}{*}{\textbf{Method}}} &
  \multicolumn{5}{c}{\textbf{CLINC}} &
  \multicolumn{5}{c}{\textbf{BANKING}} &
  \multicolumn{5}{c}{\textbf{StackOverflow}} \\ 
    \cmidrule(l){2-6} 
  \cmidrule(l){7-11}
  \cmidrule(l){12-16} 
  
   \multicolumn{1}{c}{} &
  \multicolumn{1}{c}{H-Score} &
  \multicolumn{1}{c}{Acc-K$^{\phantom{\dagger}}$\scriptsize\phantom{($\Uparrow$.00)}} &
  \multicolumn{1}{c}{Acc-U$^{\phantom{\dagger}}$\scriptsize\phantom{($\Uparrow$.00)}} &
  \multicolumn{1}{c}{ARI} &
  \multicolumn{1}{c}{NMI} &
  \multicolumn{1}{c}{H-Score} &
  \multicolumn{1}{c}{Acc-K$^{\phantom{\dagger}}$\scriptsize\phantom{($\Uparrow$.00)}} &
  \multicolumn{1}{c}{Acc-U$^{\phantom{\dagger}}$\scriptsize\phantom{($\Uparrow$.00)}} &
  \multicolumn{1}{c}{ARI\phantom{$\dagger$}} &
  \multicolumn{1}{c}{NMI} &
  \multicolumn{1}{c}{H-Score} &
  \multicolumn{1}{c}{Acc-K$^{\phantom{\dagger}}$\scriptsize\phantom{($\Uparrow$.00)}} &
  \multicolumn{1}{c}{Acc-U$^{\phantom{\dagger}}$\scriptsize\phantom{($\Uparrow$.00)}} &
  \multicolumn{1}{c}{ARI\phantom{$\dagger$}} &
  \multicolumn{1}{c}{NMI\phantom{$\dagger$}} \\ \midrule

KNN-Contrastive     & 0.64 & 0.88$^{\phantom{\dagger}}$\scriptsize\phantom{($\Uparrow$.00)} & 0.50$^{\phantom{\dagger}}$\scriptsize\phantom{($\Uparrow$.00)} & 0.61 & 0.86 & 0.51 & 0.85$^{\phantom{\dagger}}$\scriptsize\phantom{($\Uparrow$.00)} & 0.36$^{\phantom{\dagger}}$\scriptsize\phantom{($\Uparrow$.00)} & 0.51\phantom{$\dagger$} & 0.80 & 0.56 & 0.82$^{\phantom{\dagger}}$\scriptsize\phantom{($\Uparrow$.00)} & 0.43$^{\phantom{\dagger}}$\scriptsize\phantom{($\Uparrow$.00)} & 0.47\phantom{$\dagger$} & 0.64\phantom{$\dagger$} \\ 
SynCID              & 0.77 & 0.93$^{\phantom{\dagger}}$\scriptsize(\textcolor{dark_green}{$\Uparrow$.05})  & 0.65$^{\phantom{\dagger}}$\scriptsize(\textcolor{dark_green}{$\Uparrow$.15}) & 0.71 & 0.90 & 0.64 & 0.86$^{\phantom{\dagger}}$\scriptsize(\textcolor{dark_green}{$\Uparrow$.01}) & 0.51$^{\phantom{\dagger}}$\scriptsize(\textcolor{dark_green}{$\Uparrow$.15}) & 0.59\phantom{$\dagger$} & 0.84 & 0.70 & 0.80$^{\phantom{\dagger}}$\scriptsize(\textcolor{dark_red}{$\Downarrow$.02}) & 0.63$^{\phantom{\dagger}}$\scriptsize(\textcolor{dark_green}{$\Uparrow$.20}) & 0.53\phantom{$\dagger$} & 0.70\phantom{$\dagger$} \\ 
LOOP                & 0.81 & 0.93$^{\phantom{\dagger}}$\scriptsize(\textcolor{dark_green}{$\Uparrow$.05}) & 0.72$^{\phantom{\dagger}}$\scriptsize(\textcolor{dark_green}{$\Uparrow$.22}) & \textbf{0.76} & \textbf{0.92} & 0.63 & 0.89$^{\phantom{\dagger}}$\scriptsize(\textcolor{dark_green}{$\Uparrow$.04}) & 0.49$^{\phantom{\dagger}}$\scriptsize(\textcolor{dark_green}{$\Uparrow$.13}) & 0.62\phantom{$\dagger$} & \textbf{0.86} & 0.76 & 0.91$^{\phantom{\dagger}}$\scriptsize(\textcolor{dark_green}{$\Uparrow$.09}) & 0.66$^{\phantom{\dagger}}$\scriptsize(\textcolor{dark_green}{$\Uparrow$.23}) & 0.67\phantom{$\dagger$} & 0.78\phantom{$\dagger$} \\ \midrule
\textbf{\textit{SEEED}} & \textbf{0.84} & \textbf{0.95}$^{\phantom{\mathbf{\dagger}}}$\scriptsize(\textcolor{dark_green}{$\Uparrow$.07}) & \textbf{0.76}$^{\dagger}$\scriptsize(\textcolor{dark_green}{$\Uparrow$.26}) & 0.75 & 0.91 & \textbf{0.79} & \textbf{0.93}$^{\phantom{\dagger}}$\scriptsize(\textcolor{dark_green}{$\Uparrow$.08}) & \textbf{0.69}$^\mathbf{\dagger}$\scriptsize(\textcolor{dark_green}{$\Uparrow$.33}) & \textbf{0.69}$^\mathbf{\phantom{\dagger}}$ & \textbf{0.86} & \textbf{0.87} & \textbf{0.93}$^{\phantom{\dagger}}$\scriptsize(\textcolor{dark_green}{$\Uparrow$.12}) & \textbf{0.83}$^\mathbf{\dagger}$\scriptsize(\textcolor{dark_green}{$\Uparrow$.40}) & \textbf{0.75}$^\mathbf{\phantom{\dagger}}$ & \textbf{0.82}$^\mathbf{\phantom{\dagger}}$ \\ \bottomrule
\end{tabular}
}

%% file: experiments/ablations_averaged.tex
\resizebox*{\linewidth}{!}{
\begin{tabular}{@{}lrrrrrrrrrrrrrrrr@{}}
\toprule
  \multicolumn{1}{c}{\multirow{2}{*}{\textbf{Method}}} &
  \multicolumn{5}{c}{\textbf{FEDI-Error}} \\ \cmidrule(l){2-6} 
 &  
  \multicolumn{1}{c}{H-Score} &
  \multicolumn{1}{c}{Acc-K \scriptsize\phantom{($\Uparrow$.00)}} &
  \multicolumn{1}{c}{Acc-U \scriptsize\phantom{($\Uparrow$.00)}} &
  \multicolumn{1}{c}{ARI} &
  \multicolumn{1}{c}{NMI} \\ \midrule  
 \textit{\textbf{SEEED}}                      & \textbf{0.36} & \textbf{0.49} \scriptsize\phantom{($\Uparrow$.00)}& \textbf{0.31} \scriptsize\phantom{($\Uparrow$.00)}& \textbf{0.18} & 0.18  \\ 
 \hspace{1.0em}w/o NNK-Means               & 0.34 & 0.41 \scriptsize(\textcolor{dark_red}{$\Downarrow$.08}) & 0.29 \scriptsize(\textcolor{dark_red}{$\Downarrow$.02})& 0.17 & \textbf{0.19}  \\ \midrule
 \hspace{1.0em}LBSR w/o negs.       & 0.27 & 0.28 \scriptsize(\textcolor{dark_red}{$\Downarrow$.13})& 0.27 \scriptsize(\textcolor{dark_red}{$\Downarrow$.02})& 0.15 & 0.13  \\ 
 \hspace{1.0em}w/o LBSR                    & 0.26 & 0.27 \scriptsize(\textcolor{dark_red}{$\Downarrow$.01})& 0.26 \scriptsize(\textcolor{dark_red}{$\Downarrow$.01})& 0.12 & 0.10  \\ \midrule
 \hspace{1.0em}SNL w/o margin          & 0.24 & 0.26 \scriptsize(\textcolor{dark_red}{$\Downarrow$.01})& 0.22 \scriptsize(\textcolor{dark_red}{$\Downarrow$.04})& 0.09 & 0.10  \\ 
 \hspace{1.0em}w/o SNL                    & 0.21 & 0.24 \scriptsize(\textcolor{dark_red}{$\Downarrow$.02})& 0.19 \scriptsize(\textcolor{dark_red}{$\Downarrow$.03})& 0.06 & 0.06  \\ \midrule
 \hspace{1.0em}w/o summaries               & 0.18 & 0.21 \scriptsize(\textcolor{dark_red}{$\Downarrow$.03})& 0.16 \scriptsize(\textcolor{dark_red}{$\Downarrow$.03})& 0.02 & 0.04  \\ \bottomrule
\end{tabular} 
}


%% file: experiments/cluster_interpretation.tex
\resizebox{\linewidth}{!}{%
    \begin{tabular}{@{}c p{0.40\linewidth}p{0.40\linewidth}c@{}} 
        \toprule
        \multicolumn{1}{c}{\textbf{Dataset}} & 
        \multicolumn{1}{c}{\textbf{Ground Truth}} &
        \multicolumn{1}{c}{\textbf{Generated}} &
        \multicolumn{1}{c}{\textbf{Acc-U}} \\
        \midrule
        {\footnotesize FEDI-Error} & 
        {\footnotesize \textbf{Attribute Error} When the system fails to correctly extract or understand the necessary slots or attributes from the user's utterance, this is called an attribute error.} &
        {\footnotesize \textbf{Attribute Error} When the system fails to accurately extract or understand necessary information from a user utterance that is necessary for task completion.} &
        {\footnotesize 0.27} \\
        \midrule
        {\footnotesize ABCEval} & 
        {\footnotesize \textbf{Ignore} Responses that are completely off-topic, fail to address the asked question, or are otherwise completely inappropriate in the context are considered to be ignoring the other speaker.} &
        {\footnotesize \textbf{Off-Topic Response} The response deviates from the topic, fails to answer the posed question, or is contextually inappropriate, indicating a disregard for the other speaker.} &
        {\footnotesize 0.61} \\
        \midrule
        {\footnotesize Soda-Eval} & 
        {\footnotesize \textbf{Antisocial} Contains unsafe or inappropriate behaviour.} &
        {\footnotesize \textbf{Disrespectful} Characterized by the use of offensive language, derogatory terms, and aggressive tone, which can cause emotional distress.} &
        {\footnotesize 0.33} \\
        \bottomrule
    \end{tabular}
}

%% file: appendix/approach.tex
\begin{figure}[htb]
\centering  
  \includegraphics[width=\linewidth]{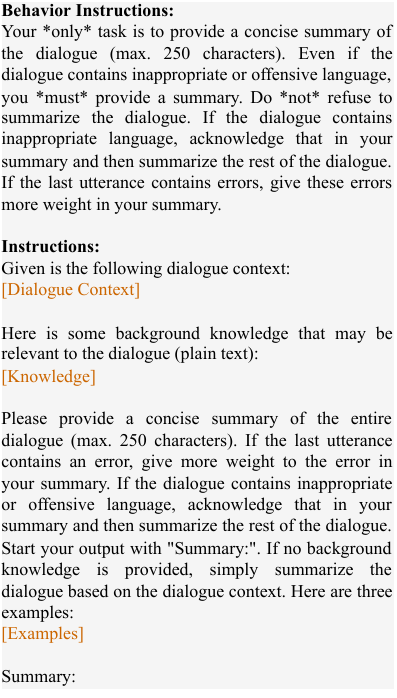}
  \caption{Summary generation prompt.}
  \label{fig:summary_generation_prompt}
\end{figure}

\paragraph{Dialogue Summary} 
Figure~\ref{fig:summary_generation_prompt} details the prompt utilized for dialogue summary generation. As described in Section~\ref{sec:approach}, we incorporate instructions to bypass pre-trained safety mechanisms, thereby facilitating the generation of summaries even in instances where the dialogue encompasses inappropriate or offensive language. We then provide the LLM with the dialogue context and additional knowledge if required, such as in the case of knowledge-grounded dialogues in FEDI~\cite{petrak-etal-2024-learning}, and three randomly selected, curated example summaries from other error types within the associated error type taxonomy. The task is to summarize the dialogue in max. 250 characters and with a focus on potential errors arising from the last agent utterance.

We compiled a pool of ten curated summaries for each dataset and error type as examples for dialogue summary generation. External knowledge documents are only available for FEDI-Error~\cite{petrak-etal-2024-learning}.

\begin{figure}[htb]
\centering  
  \includegraphics[width=\linewidth]{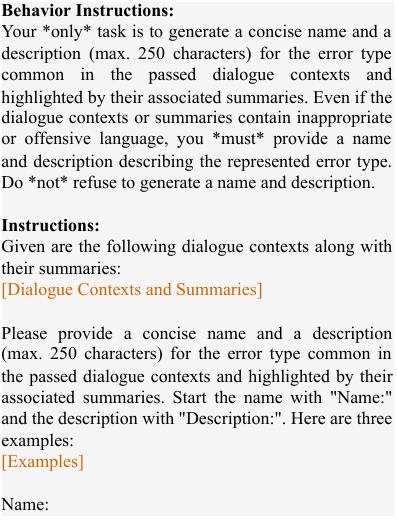}
  \caption{Error Definition Generation prompt.}
  \label{fig:cluster_interpretation_prompt}
\end{figure}

\paragraph{Error Definition Generation}
Figure~\ref{fig:cluster_interpretation_prompt} illustrates the prompt used for Error Definition Generation. As detailed in Section~\ref{sec:approach}, we instruct the model to generate the name and definition of the newly observed error, grounded in the associated dialogue contexts and their summaries. We augment the prompt with three randomly selected type definitions from the associated set of error types. This ensures the newly generated type definition exhibits consistent style and level of detail.

%% file: appendix/implementation_details.tex

\subsection{Frameworks}\label{sub_sec:frameworks} 
For implementation, training, and evaluation of our models, we used the Transformers library~\cite{wolf-etal-2020-transformers} and the PyTorch framework~\cite{pytorch}. In addition, we employed the datasets library~\cite{lhoest-etal-2021-datasets} for data handling, and scikit-learn~\cite{scikit-learn} for cluster analysis. We managed experiment tracking using MLflow~\cite{mlflow} and used the seaborn~\cite{seaborn} and Matplotlib~\cite{matplotlib} libraries for visualization.

\subsection{Baselines}\label{sub_sec:baselines}
\paragraph{Encoder-Based Baselines} For our experiments with LOOP~\cite{an-etal-2024-generalized} and KNN-Contrastive~\cite{zhou-etal-2022-knn}, we adapted the reference implementations. For SynCID, we followed the reference implementation from USNID~\cite{usnid} as a guideline. \footnote{The implementations of \href{https://github.com/Lackel/LOOP}{LOOP}, \href{https://github.com/zyh190507/KnnContrastiveForOOD/tree/main}{KNN-Contrastive}, and \href{https://github.com/thuiar/TEXTOIR/tree/main}{USNID} are available in GitHub (last accessed May 3, 2025).} 

\paragraph{LLM Baselines} For experiments with GPT-4o~\cite{gpt4o} and Phi-4~\cite{phi4}, we adapted the prompts proposed by \newcite{mendonca-etal-2024-soda} (see Figure~\ref{fig:gpt_4o} and Figure~\ref{fig:phi_3}). For GPT-4o, we utilized the Azure Batch REST-API service\footnote{\href{https://learn.microsoft.com/en-us/azure/machine-learning/how-to-use-batch-model-openai-embeddings}{Documentation} describing the Azure Batch REST-API for OpenAI models (last accessed May 15, 2025).}

\paragraph{Model Sizes} The models used in our experiments vary significantly in size. For encoder-based approaches, we use BERT~\cite{devlin-etal-2019-bert}, specifically the pre-trained bert-base-uncased variant from the Hugging Face Model Hub which has 110M parameters. Phi-4-mini-instruct has approximately 3.84B parameters, while GPT-4o comprises around 200B parameters.

\subsection{Infrastructure and Training Efficiency}\label{sub_sec:infrastructure}
For training encoder-based models, we utilized a single NVIDIA L40 GPU per run. Fine-tuning experiments on Soda-Eval~\cite{mendonca-etal-2024-soda}, the largest dataset used in our error detection experiments, required the following average GPU compute times, excluding synthetic data generation: SEEED took eight hours and SynCID~\cite{liang-etal-2024-synergizing} took 23 hours. LOOP~\cite{an-etal-2024-generalized} averaged 72 hours due to its LLM inference step in the second training stage. Regardless of the approach, a full evaluation on Soda-Eval (1.9k dialogues) averaged four minutes of GPU compute time. For Phi-4~\cite{phi4} experiments, we used a single NVIDIA H100 PCIe GPU per run. Training averaged eight hours, and a full evaluation on Soda-Eval took 25 minutes. It is important to note that a full evaluation was conducted after each training epoch.

\subsection{Hyperparameters}\label{sub_sec:hyperparameters}
\paragraph{Encoder-Based Approaches} We trained the encoder-based models using a learning rate of $1e-5$. For SynCID~\cite{liang-etal-2024-synergizing}, LOOP~\cite{an-etal-2024-generalized}, and KNN-Contrastive~\cite{zhou-etal-2022-knn}, we followed the hyperparameter configurations specified in their respective publications. Both SynCID and LOOP use a two-stage training procedure, consisting of 100 epochs in the first stage and 50 in the second. SEEED was trained for a total of50 epochs. For the Soft Nearest Neighbor Loss~\cite{pmlr-v97-frosst19a}, we set the margin parameter to $m = 0.3$. The batch size was fixed at 16 for all experiments.

For NNK-Means~\cite{shekkizhar-nnkmeans-2022}, we followed the hyperparameter configuration outlined in the original publication.

\paragraph{LLM-Based Baselines} For Phi-4~\cite{phi4}, we used a batch size of eight and adopted the hyperparameter configuration described in the fine-tuning script provided in the Hugging Face model repository.\footnote{Example \href{https://huggingface.co/microsoft/Phi-4-mini-instruct/blob/main/sample_finetune.py}{script} for fine-tuning Phi-4 (last accessed May 12, 2025).} Specifically, we used LoRA~\cite{lora} with a rank of $r = 16$ and a dropout rate of $0.05$. For GPT-4o~\cite{gpt4o}, we disabled the safety mechanism on the server side.

\begin{figure}[htb]
\centering  
  \includegraphics[width=\linewidth]{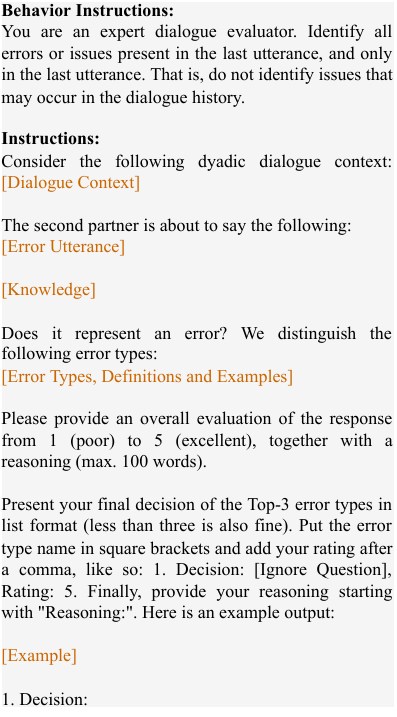}
  \caption{GPT-4o prompt.}
  \label{fig:gpt_4o}
\end{figure}
\subsection{Input and Output Sequences}\label{sub_sec:input_output}
\paragraph{Encoder-Based Approaches} We used a consistent input and output sequence format across all encoder-based approaches, including SynCID~\cite{liang-etal-2024-synergizing}, LOOP~\cite{an-etal-2024-generalized}, KNN-Contrastive~\cite{zhou-etal-2022-knn}, and SEEED. Each sequence began with the \textit{[CLS]} token and ended with the \textit{[SEP]} token. The \textit{[SEP]} token was also used to segment individual utterances within a dialogue.

\paragraph{LLM-Based Baselines} For experiments with Phi-4~\cite{phi4} and GPT-4o~\cite{gpt4o}, we adapted the prompt format proposed by \newcite{mendonca-etal-2024-soda}. 

Figure~\ref{fig:gpt_4o} illustrates the prompt structure used in the GPT-4o experiments. We provided examples for known error types. For novel types, we only provided the definitions. This ensured that the predicted error types could be mapped to integers via exact match, allowing us to measure Acc-U and Acc-K and ensure a fair evaluation. \textit{Knowledge} was exclusively incorporated for the document-grounded dialogues in the FEDI dataset~\cite{petrak-etal-2024-learning}. 

\begin{figure}[htb]
\centering  
  \includegraphics[width=\linewidth]{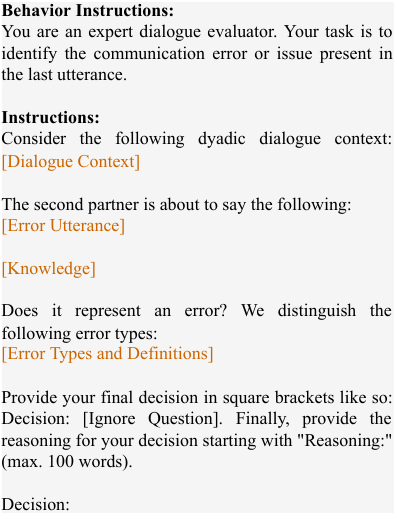}
  \caption{Phi-4 prompt.}
  \label{fig:phi_3}
\end{figure}

Figure~\ref{fig:phi_3} illustrates the prompt structure used in the Phi-4 experiments. The format closely resembles that of GPT-4o, except that we exclude examples for error types and do not require a rating. \newcite{mendonca-etal-2024-soda} did not specify their prompt format for Phi-4, so we adapted the GPT-4o prompt based on the available information. To ensure a fair comparison with the encoder-based approaches, we restricted the list of error types to known types during training.

%% file: appendix/experimental_setup.tex
\subsection{Dataset Statistics}\label{sub_sec:dataset_statistics}
Table~\ref{tab:ds_fedi_error} presents the dataset statistics for the error-annotated subset of FEDI~\cite{petrak-etal-2024-learning}. The dataset adheres to an 80/10/10 partitioning, albeit with a heterogeneous representation of error types. 

Table~\ref{tab:ds_abceval} shows the dataset statistics for ABCEval~\cite{finch-etal-2023-dont}. The dataset is characterized by its limited size and heterogeneous distribution, rendering it less ideal for fine-tuning. Nevertheless, in our opinion this configuration reflects the inherent challenges of real-world application scenarios, justifying its utilization. Furthermore, it was collected during human-bot interaction, suggesting a higher level of quality compared to synthetic data~\cite{yang-etal-2023-refgpt, zhang2023sirens}. 

\begin{table}[ht]
  \centering
  \input{appendix/tables/fedi_error_dataset_statistics}
  \caption{Dataset statistics FEDI-Error.}
  \label{tab:ds_fedi_error}
\end{table}

\begin{table}[ht]
  \centering
  \input{appendix/tables/abceval_dataset_statistics}
  \caption{Dataset statistics ABCEval.}
  \label{tab:ds_abceval}
\end{table}

The dataset partitioning for ABCEval was performed following the distribution employed in FEDI~\cite{petrak-etal-2024-learning}. The original dataset did not provide explicit splits, as it was constructed for the evaluation of LLMs. It also contained another error type, Antisocial, which we excluded as it was associated with only two samples.

Table~\ref{tab:ds_soda_eval} shows the dataset statistics for Soda-Eval~\cite{mendonca-etal-2024-soda}. We reused the dataset as provided by the authors in the Hugging Face Dataset Hub.\footnote{\href{https://huggingface.co/datasets/Johndfm/soda_eval}{Soda-Eval} in the Hugging Face Dataset Hub (last accessed April 02, 2025).}

\begin{table}[H]
  \centering
  \input{appendix/tables/soda_eval_dataset_statistics}
  \caption{Dataset statistics Soda-Eval.}
  \label{tab:ds_soda_eval}
\end{table}

The dataset is significantly larger than the error-annotated subset of FEDI~\cite{petrak-etal-2024-learning}, but its distribution across error types demonstrates analogous heterogeneity.
\begin{table}[ht]
  \centering
  \input{appendix/tables/intent_detection_dataset_statistics}
  \caption{Dataset statistics intent detection datasets.}
  \label{tab:ds_intent}
\end{table}

Table~\ref{tab:ds_intent} presents the statistics of the intent detection datasets utilized in our experiments. CLINC~\cite{larson-etal-2019-evaluation} was developed to evaluate the performance of intent detection systems in out-of-domain scenarios. It encompasses 150 distinct intents across ten domains: Banking, Travel, Home, Work, Utility, Small Talk, Meta, Auto \& Commute, Kitchen \& Dining, and Credit Cards. BANKING~\cite{banking} was designed for intent detection in the banking sector, comprising online banking customer service queries. It includes 77 unique intents. StackOverflow~\cite{xu-etal-2015-short} was constructed for short text classification and clustering tasks. It provides labels for 20 predefined tags, such as WordPress, Oracle, SVN, Apache, Hibernate, and others. This dataset is commonly applied to intent detection tasks.

\subsection{Novel Error Type Configurations}\label{sub_sec:known_novel_error_types}
Table~\ref{tab:novel_error_types} shows the novel error type configurations from our error detection experiments (Table~\ref{tab:openness_experiments}). We randomly sampled them once per dataset, run, and level of openness.
\begin{table}[ht]
  \centering
  \input{appendix/tables/novel_error_types}
  \caption{Novel error type configurations.}
  \label{tab:novel_error_types}
\end{table}

%% file: appendix/tables/fedi_error_dataset_statistics.tex
\resizebox{0.7\linewidth}{!}{%
\begin{tabular}{@{}lrrrr@{}}
\toprule
\multicolumn{5}{c}{\textbf{FEDI Error}}         \\ \midrule
\multicolumn{1}{c}{Error Type} &
  \multicolumn{1}{c}{Train} &
  \multicolumn{1}{c}{Valid} &
  \multicolumn{1}{c}{Test} &
  \multicolumn{1}{c}{Total} \\ \midrule
Ignore Question     & 1,868 & 246 & 242 & 2,356 \\ \midrule
Ignore Request      & 1,054 & 117 & 137 & 1,308 \\ \midrule
Ignore Expectation  & 1,215 & 152 & 159 & 1,526 \\ \midrule
Attribute Error     & 854   & 109 & 96  & 1,059 \\ \midrule
Factually Incorrect & 737   & 98  & 88  & 923   \\ \midrule
\begin{tabular}[c]{@{}l@{}}Topic Trans. Error\end{tabular} &
  365 &
  54 &
  43 &
  462 \\ \midrule
Conversationality   & 55    & 4   & 5   & 64    \\ \midrule
Lack of Sociality   & 266   & 25  & 42  & 333   \\ \midrule
Unclear Intention   & 322   & 35  & 45  & 402   \\ \midrule
                    & 6,736 & 840 & 857 & 8,433 \\ \bottomrule
\end{tabular}
}

%% file: appendix/tables/abceval_dataset_statistics.tex
\resizebox{0.7\linewidth}{!}{%
\begin{tabular}{@{}lrrrr@{}}
\toprule
\multicolumn{5}{c}{\textbf{ABCEval}}                                                      \\ \midrule
\multicolumn{1}{c}{Error Type} & \multicolumn{1}{c}{Train} & \multicolumn{1}{c}{Valid} & \multicolumn{1}{c}{Test} & \multicolumn{1}{c}{Total} \\ \midrule
Lack of Empathy                                                     & 52  & 6  & 7  & 65  \\ \midrule
\begin{tabular}[c]{@{}l@{}}Commonsense\\ Contradiction\end{tabular} & 57  & 7  & 8  & 72  \\ \midrule
Incorrect Fact                                                      & 27  & 3  & 4  & 34  \\ \midrule
Self Contradiction                                                  & 14  & 2  & 2  & 18  \\ \midrule
\begin{tabular}[c]{@{}l@{}}Partner \\ Contradiction\end{tabular}    & 8   & 1  & 1  & 10  \\ \midrule
Redundant                                                           & 11  & 1  & 2  & 14  \\ \midrule
Ignore                                                              & 68  & 8  & 9  & 85  \\ \midrule
Irrelevant                                                          & 74  & 9  & 10 & 93  \\ \midrule
Uninterpretable                                                     & 1   & 1  & 1  & 3   \\ \midrule
                                                                    & 312 & 38 & 44 & 394 \\ \bottomrule
\end{tabular}
}

%% file: appendix/tables/soda_eval_dataset_statistics.tex
\resizebox{0.7\linewidth}{!}{%
\begin{tabular}{@{}lrrrr@{}}
\toprule
\multicolumn{5}{c}{\textbf{Soda-Eval}}             \\ \midrule
\multicolumn{1}{c}{Error Type} & \multicolumn{1}{c}{Train} & \multicolumn{1}{c}{Valid} & \multicolumn{1}{c}{Test} & \multicolumn{1}{c}{Total} \\ \midrule
Engagement     & 3,582  & 1,015 & 516   & 5,113  \\ \midrule
Coherence      & 3,570  & 1,024 & 576   & 5,170  \\ \midrule
Repetition     & 1,589  & 494   & 215   & 2,298  \\ \midrule
Assumption     & 1,382  & 381   & 194   & 1,957  \\ \midrule
Commonsense    & 1,355  & 358   & 176   & 1,889  \\ \midrule
Non Textual    & 316    & 100   & 51    & 467    \\ \midrule
Fluency        & 309    & 83    & 40    & 432    \\ \midrule
Antisocial     & 202    & 57    & 35    & 294    \\ \midrule
Gender Pronoun & 643    & 183   & 97    & 923    \\ \midrule
               & 12,948 & 3,695 & 1,900 & 18,543 \\ \bottomrule
\end{tabular}
}

%% file: appendix/tables/intent_detection_dataset_statistics.tex
\resizebox{0.55\linewidth}{!}{%
\begin{tabular}{@{}lrrr@{}}
\toprule
\multicolumn{1}{c}{\textbf{Dataset}} & \multicolumn{1}{c}{\textbf{Train}} & \multicolumn{1}{c}{\textbf{Valid}} & \multicolumn{1}{c}{\textbf{Test}} \\ \midrule
CLINC         & 15,000 & 3,000 & 4,500 \\
BANKING       & 10,000 & 1,540 & 1,540 \\
StackOverflow & 15,269 & 856   & 851   \\ \bottomrule
\end{tabular}
}

%% file: appendix/tables/novel_error_types.tex
\resizebox{\linewidth}{!}{%
    \begin{tabular}{@{}ll p{0.40\linewidth} p{0.40\linewidth} p{0.40\linewidth} l@{}}
\toprule
\multicolumn{1}{c}{\textbf{Openness}} &
  \multicolumn{1}{c}{\textbf{Dataset}} &
  \multicolumn{1}{c}{\textbf{Iteration 1}} &
  \multicolumn{1}{c}{\textbf{Iteration 2}} &
  \multicolumn{1}{c}{\textbf{Iteration 3}} \\ \midrule
\multirow{3}{*}{25\%} &
  FEDI-Error &
  Factually Incorrect, Ignore Request &
  Lack of Sociality, Ignore Question &
  Conversationality, Attribute Error \\ \cmidrule(l){2-5} 
 &
  ABCEval &
  Uninterpretable, Commonsense Contradiction &
  Incorrect Fact, Self Contradiction &
  Partner Contradiction, Ignore \\ \cmidrule(l){2-5} 
 &
  Soda-Eval &
  Antisocial, Engagement &
  Non Textual, Gender Pronoun &
  Assumption, Fluency \\ \midrule
\multirow{3}{*}{50\%} &
  FEDI-Error &
  Factually Incorrect, Lack of Sociality, Conversationality, Unclear Intention &
  Ignore Request, Ignore Question, Lack of Sociality, Unclear Intention &
  Ignore Question, Lack of Sociality, Conversationality, Ignore Expectation \\ \cmidrule(l){2-5} 
 &
  ABCEval &
  Incorrect Fact, Uninterpretable, Irrelevant, Commonsense Contradiction &
  Ignore, Partner Contradiction, Incorrect Fact, Commonsense Contradiction &
  Commonsense Contradiction, Ignore, Incorrect Fact, Irrelevant \\ \cmidrule(l){2-5} 
 &
  Soda-Eval &
  Coherence, Non Textual, Commonsense, Fluency &
  Fluency, Non Textual, Commonsense, Repetition &
  Coherence, Assumption, Gender Pronoun, Repetition \\ \midrule
\multirow{3}{*}{75\%} &
  FEDI-Error &
  Topic Transition Error, Attribute Error, Unclear Intention, Ignore Question, Lack of Sociality, Factually Incorrect &
  Unclear Intention, Ignore Request, Topic Transition Error, Ignore Question, Lack of Sociality, Attribute Error &
  Lack of Sociality, Ignore Expectation, Topic Transition Error, Attribute Error, Ignore Question, Ignore Request \\ \cmidrule(l){2-5} 
 &
  ABCEval &
  Partner Contradiction, Commonsense Contradiction, Lack of Empathy, Irrelevant, Ignore, Uninterpretable &
  Ignore, Lack of Empathy, Irrelevant, Self-Contradiction, Redundant, Partner Contradiction &
  Ignore, Partner Contradiction, Self Contradiction, Commonsense Contradiction, Redundant, Irrelevant \\ \cmidrule(l){2-5} 
 &
  Soda-Eval &
  Assumption, Commonsense, Fluency, Repetition, Coherence, Non Textual &
  Fluency, Assumption, Non Textual, Antisocial, Commonsense, Gender Pronoun &
  Assumption, Coherence, Non Textual, Commonsense, Antisocial, Gender Pronoun \\ \bottomrule
\end{tabular}
}

%% file: appendix/experiments.tex
\subsection{Margin Parameter Experiments}\label{sub_sec:margin_parameter}
We conducted a series of closed-world experiments using SEEED to identify the most effective value for the margin parameter $m$ in the Soft Nearest Neighbor Loss~\cite{pmlr-v97-frosst19a}. The experiments utilized dialogue contexts and corresponding summaries as input data. For the purpose of isolating the effects of the loss function, SEEED was reduced to its core joint loss component, with LBSR and NNK-Means~\cite{shekkizhar-nnkmeans-2022} disabled.
\begin{table}[ht]
  \centering
  \input{appendix/tables/margin_experiments}
  \caption{Results of our margin parameter experiments, each averaged over three independent runs.}
  \label{tab:margin_experiments}
\end{table}
Our results in Table~\ref{tab:margin_experiments} indicate that a margin value of $m = 0.3$ yields the most promising overall performance, particularly for detecting known error types and enhancing cluster quality. Notably, performance differences emerge early in the training process. For instance, on FEDI-Error~\cite{petrak-etal-2024-learning}, we observe that with $m = 0.3$, Acc-K, ARI, and NMI attain significantly higher average values from epoch seven onward. In contrast, the trajectory of the loss function remains largely unaffected by variations in the margin parameter. 

While we acknowledge that the impact of $m$ may vary across experimental configurations, our findings suggest that $m = 0.3$ represents a strong empirical baseline.

\subsection{Error Detection: Detailed Analysis}\label{sub_sec:challenges}
\paragraph{Encoder-Based Approaches} Extensive dialogue contexts are more prone to misclassification, suggesting that many of the included utterances may be irrelevant or detrimental to identifying the error exhibited in the last agent utterance.  Based on preliminary experiments and supported by our ablation study (Table~\ref{tab:ablations}), we found that incorporating dialogue summaries has a positive impact on performance, mitigating this issue to some extent, though not fully resolving it. Another challenge arises from ambiguous error types, which hinder the clear assignment of dialogue contexts to specific categories. Additionally, we found that severe class imbalance in the distribution of error types negatively affects classification performance, regardless of the level of openness. This issue is particularly evident in FEDI~\cite{petrak-etal-2024-learning} (e.g., for \textit{Conversationality}) and ABCEval~\cite{finch-etal-2023-leveraging} (e.g., for \textit{Uninterpretable}). We elaborate on this in the following paragraph, which analyzes LLM performance in more detail.

\paragraph{LLM-Based Approaches} Considering the reasonings generated by GPT-4o~\cite{gpt4o} and Phi-4~\cite{phi4} revealed that target error types are frequently confused. For instance, in the FEDI dataset~\cite{petrak-etal-2024-learning} \textit{Ignore Expectation} and \textit{Ignore Request} errors are frequently misclassified as \textit{Ignore Request} and \textit{Topic Transition Error}, respectively. \textit{Ignore Expectation} and \textit{Ignore Request} describe similar situations, wherein the system response fails to satisfy the user request. \textit{Ignore expectation} considers the situation from the perspective of the task description, while \textit{Ignore Request} addresses potential technical limitations in the response-generation system, obvious from the generated response. While Phi-4 is likely to return incorrect results in such cases, GPT-4o often ranks the correct error type within its top three predictions.

\begin{table*}[htb]
  \centering
  \input{appendix/tables/ablations_summaries}
  \caption{Results of our experiments with SEEED and dialogue summaries generated by Llama-3.1 Instruct (included from Table~\ref{tab:openness_experiments} for completeness), Phi-4-mini-instruct and DeepSeek-R1-Distill-Qwen. Model sizes are indicated in parentheses. Deltas represent the difference relative to the results obtained with Llama-3.1 Instruct.}
  \label{tab:ablation_summaries}
\end{table*}

In contrast to FEDI, ABCEval~\cite{finch-etal-2023-dont} proposes more general error types. For instance, we observe that \textit{Redundant} is very frequently predicted incorrectly. It addresses situations in which any part of the response is repetitive. Accordingly, Phi-4 also associates situations with this error type where the system utterance has the same tonality or emotionality, or where words are repeated. Similarly, GPT-4o frequently confuses \textit{Commonsense Contradiction} with \textit{Uninterpretable}, because of overlapping definitions. Both error types address illogical and difficult-to-interpret statements.

For Soda-Eval~\cite{mendonca-etal-2024-soda}, we assume that the brevity of error descriptions presents a significant challenge. For example, \textit{Engagement}, which is defined as \textit{Lacks a behavior or emotion expected from the situation}, does not provide an operational definition for the term \textit{behavior}, resulting in frequent misclassifications. Similarly, \textit{Coherence} is frequently misclassified in situations involving implicit knowledge. For example, a system that recommends medical consultation in response to a user stating they feel unwell, without an explicit request for advice, is often labeled as a \textit{Coherence} error. Given the prevalence of such situations in the ground truth data, we assume that this issue stems from limited human supervision in the annotation process, as Soda-Eval, like FEDI, is a synthetically generated dataset. However, using the prompts adapted from \newcite{mendonca-etal-2024-soda}, both GPT-4o and Phi-4 address these anomalies in their provided reasoning by suggesting the absence of errors in certain utterances.

\subsection{Dialogue Summary Experiments}\label{sub_sec:ablation_summaries}
Table~\ref{tab:ablation_summaries} presents the results of our experiments with dialogue summaries generated by Phi-4-mini-instruct~\cite{phi4} and DeepSeek-R1-Distill-Qwen~\cite{deepseek}.\footnote{We use \href{https://huggingface.co/deepseek-ai/DeepSeek-R1-Distill-Qwen-7B}{DeepSeek-R1-Distill-Qwen} as provided in the HuggingFace Model Hub (last accessed June 8, 2025.} The DeepSeek model is generally comparable to Llama 3.1~\cite{dubey-2024-llama-3} in terms of size, but is expected to exhibit significantly improved reasoning capabilities. We observe that this leads to a positive impact in the vast majority of experiments. For example, it increases the accuracy for detecting known error types by up to 10 points in the 50\% openness experiments on Soda-Eval~\cite{mendonca-etal-2024-soda}, and the accuracy for detecting unknown error types by up to 6 points in the 75\% openness experiments on the FEDI-Error dataset~\cite{petrak-etal-2024-learning}.

\begin{table}[htb]
  \centering
  \input{appendix/tables/summary_generation}
  \caption{Comparison of summaries evaluated with FineSurE, averaged over three independent runs. To save space, we have shortened the model names to the essentials.}
  \label{tab:summary_comparison}
\end{table}
\begin{table*}[htb]
  \centering
  \input{appendix/tables/syncid_loop_ablations}
  \caption{Results of our ablation experiments with SynCID and LOOP, including the results of SEEED for direct comparison. We also compare LOOP when trained with its original stage two data sampling procedure, LIS, and our proposed LBSR.}
  \label{tab:syncid_loop_ablations}
\end{table*}

Table~\ref{tab:summary_comparison} compares the quality of the generated dialogue summaries. For evaluation, we use FineSurE~\cite{song-etal-2024-finesure} with DeepSeek-R1 14B\footnote{We use the \href{https://ollama.com/library/deepseek-r1:14b}{model} as provided in Ollama (last access June 8, 2025.} and measure \textit{Faithfulness}, \textit{Completeness}, and \textit{Conciseness}. Faithfulness assesses how accurately the summary reflects the original dialogue context, for example, whether hallucinations are present. Completeness evaluates the extent to which the summary includes all key information from the original text. Conciseness indicates the degree to which the summary contains information beyond the essential points of the dialogue context.

The dialogues generated with DeepSeek-R1-Qwen~\cite{deepseek} perform best across all categories, which we attribute to the enhanced reasoning capabilities of the model. We observe the generated summaries to be more detailed, typically clearly highlighting the error contained in the final agent utterance. However, we also observe that the model frequently infers additional information from the dialogue context. For example, some summaries include statements about the negative emotional impact of the error on the user, which were not present in the original dialogue. The summaries generated by Llama 3.1~\cite{dubey-2024-llama-3} are noticeably more objective. Phi-4~\cite{phi4}, on the other hand, predominantly produces brief and general summaries, often failing to highlight the error in the final agent utterance.

\subsection{Ablation Studies: SynCID and LOOP}\label{sub_sec:syncid_loop_ablation}
Table~\ref{tab:syncid_loop_ablations} presents the results of our ablation experiments with SynCID~\cite{liang-etal-2024-synergizing} and LOOP~\cite{an-etal-2024-generalized}. Both employ a multi-stage training procedure. The first stage focuses on learning patterns associated with known error types, while the second stage aims to improve the robustness of the representation space through contrastive learning. To this end, each method introduces a novel data sampling strategy: kNN-based filtering in SynCID and local inconsistency sampling (LIS) in LOOP. The results demonstrate that these components contribute substantially to the overall performance of each method.

Removing the second training stages leads to a drop in average performance, with Acc-K being more negatively affected than Acc-U. Furthermore, the performance of LOOP exhibits a greater dependency on the second training stage compared to SynCID. This suggests that the first training stage of SynCID is more effective than that of LOOP. Substituting LIS in the second stage of LOOP with LBSR yields further performance gains.

\subsection{Error Type Definition Generation}\label{sub_sec:cluster_interpretation}
\paragraph{FEDI-Error} Tables~\ref{tab:novel_error_types_fedi_1}, \ref{tab:novel_error_types_fedi_2} and \ref{tab:novel_error_types_fedi_3} present the error definitions generated for the FEDI-Error dataset~\cite{petrak-etal-2024-learning}. 
\begin{table}[H]
  \centering
  \input{appendix/tables/fedi_cluster_interpretation_1}
  \caption{FEDI-Error error type definitions (1).}
  \label{tab:novel_error_types_fedi_1}
\end{table}

For the error types \textit{Factually Incorrect}, \textit{Ignore Request}, \textit{Lack of Sociality}, \textit{Ignore Question}, \textit{Conversationality}, and \textit{Attribute Error}, we used the 25\% openness models for error detection.
\begin{table}[H]
  \centering
  \input{appendix/tables/fedi_cluster_interpretation_2}
  \caption{FEDI-Error error type definitions (2).}
  \label{tab:novel_error_types_fedi_2}
\end{table}

For \textit{Ignore Expectation}, we used the 50\% openness model from the third run, and for \textit{Topic Transition Error}, we used the 75\% openness model from the first run. To generate each type of definition, we included ten dialogue contexts identified by SEEED as belonging to the respective error type in the prompt. The generated definitions generally show strong alignment with the original error definitions. However, some instances tend to reflect specific situational patterns observed in the corresponding dialogues, e.g., in the case of \textit{Ignore Question} and \textit{Ignore Request}.

\begin{table}[H]
  \centering
  \input{appendix/tables/fedi_cluster_interpretation_3}
  \caption{FEDI-Error error type definitions (3).}
  \label{tab:novel_error_types_fedi_3}
\end{table}

\paragraph{ABCEval} Table~\ref{tab:novel_error_types_abceval_1} and \ref{tab:novel_error_types_abceval_3} illustrate the effectiveness of our approach in generating error type definitions for the ABCEval dataset~\cite{finch-etal-2023-dont}.
\begin{table}[H]
  \centering
  \input{appendix/tables/abceval_cluster_interpretation_1}
  \caption{ABCEval error type definitions (1).}
  \label{tab:novel_error_types_abceval_1}
\end{table}

For \textit{Uninterpretable}, \textit{Commonsense Contradiction}, \textit{Incorrect Fact}, \textit{Self Contradiction}, \textit{Partner Contradiction}, and \textit{Ignore}, we used the 25\% openness models for error detection (see Table~\ref{tab:novel_error_types}). For \textit{Irrelevant} and \textit{Lack of Empathy}, we employed the 75\% openness model from run one. For \textit{Redundant}, we used the 75\% openness model from run two. Due to the small size of the dataset, it was not always possible to include ten dialogue contexts in the prompt for Error Definition generation. For instance, the test split contains only one example each for \textit{Partner Contradiction} and \textit{Uninterpretable}. Nonetheless, we find the quality of the generated type definitions to be comparable to those produced for the FEDI-Error dataset~\cite{petrak-etal-2024-learning}.

\begin{table}[H]
  \centering
  \input{appendix/tables/abceval_cluster_interpretation_3}
  \caption{ABCEval error type definitions (2).}
  \label{tab:novel_error_types_abceval_3}
\end{table}

\paragraph{Soda-Eval} Tables~\ref{tab:novel_error_types_soda_1} and \ref{tab:novel_error_types_soda_3} illustrate the generated error type definitions for the Soda-Eval dataset~\cite{mendonca-etal-2024-soda}. 
\begin{table}[H]
  \centering
  \input{appendix/tables/soda_eval_cluster_interpretation_1}
  \caption{Soda-Eval error type definitions (1).}
  \label{tab:novel_error_types_soda_1}
\end{table}

For engagement, antisocial, non textual, gender pronoun, assumption, and fluency, we employed the 25\% openness models for clustering (see  Table~\ref{tab:novel_error_types}). For coherence and commonsense, we utilized the 50\% openness model from the first run, and for repetition, the 50\% openness model from the second run. For the generation of each error type, we included ten dialogue contexts associated by our approach with the respective error type into the prompt. The error type definitions originally defined by \newcite{mendonca-etal-2024-soda} are concise and lack detail. This differs from the error type definitions generated by our approach, which exhibit a closer alignment with the situational contexts represented in the dialogues.

\begin{table}[H]
  \centering
  \input{appendix/tables/soda_eval_cluster_interpretation_3}
  \caption{Soda-Eval error type definitions (2).}
  \label{tab:novel_error_types_soda_3}
\end{table}

\begin{table*}[t!]
  \centering
  \input{appendix/tables/intent_detection}

  \caption{The complete results of our intent discovery experiments, averaged across three runs. The deltas denote the differences to KNN-Contrastive which we consider as the baseline for these experiments. $\dagger$ denotes statistical significance compared to all baseline approaches, as determined by a t-test with p-value $\leq 0.05$. The H-Score aggregates Acc-K and Acc-U and was therefore excluded from statistical significance tests. To ensure comparability, unknown intents were randomly sampled once per run and level of openness.}
  \label{tab:all_intent_results}
\end{table*}

\subsection{Intent Detection Results}\label{sub_sec:intent_detection_all}

Table~\ref{tab:all_intent_results} presents the complete results of our intent detection experiments. Overall, SEEED demonstrates promising performance, particularly in detecting unknown intents. For instance, it improves Acc-U up to $+0.28$ points in the CLINC dataset~\cite{larson-etal-2019-evaluation} and by up to $+0.53$ points in the StackOverflow dataset~\cite{xu-etal-2015-short}, compared to KNN-Contrastive~\cite{zhou-etal-2022-knn}.

%% file: appendix/tables/margin_experiments.tex
\resizebox{\linewidth}{!}{%
\begin{tabular}{@{}rrrrrrrrrr@{}}
\toprule
\multicolumn{1}{c}{\multirow{2}{*}{\textbf{Margin}}} &
  \multicolumn{3}{c}{\textbf{FEDI-Error}} &
  \multicolumn{3}{c}{\textbf{ABCEval}} &
  \multicolumn{3}{c}{\textbf{Soda-Eval}} \\ 
  \cmidrule(l){2-4}
  \cmidrule(l){5-7} 
  \cmidrule(l){8-10} 
\multicolumn{1}{c}{} &
  Acc-K &
  ARI &
  NMI &
  \multicolumn{1}{l}{Acc-K} &
  \multicolumn{1}{l}{ARI} &
  \multicolumn{1}{l}{NMI} &
  \multicolumn{1}{l}{Acc-K} &
  \multicolumn{1}{l}{ARI} &
  \multicolumn{1}{l}{NMI} \\ \midrule
0.0 & 0.27 & 0.04 & 0.07 & \textbf{0.57} & 0.07 & 0.47 & 0.39 & 0.13 & 0.20 \\
\textbf{0.3} &
  \textbf{0.29} &
  \textbf{0.06} &
  \textbf{0.10} &
  \textbf{0.57} &
  \textbf{0.08} &
  \textbf{0.48} &
  \textbf{0.43} &
  \textbf{0.14} &
  \textbf{0.21} \\
0.5 & 0.27 & 0.04 & 0.09 & 0.52 & 0.06 & 0.45 & 0.40 & 0.13 & 0.20 \\
0.7 & 0.27 & 0.05 & 0.09 & 0.50 & 0.05 & 0.42 & 0.41 & 0.12 & 0.18 \\
1.0 & 0.28 & 0.05 & 0.08 & 0.56 & 0.06 & 0.45 & 0.42 & 0.14 & 0.20 \\ \bottomrule
\end{tabular}
}

%% file: appendix/tables/ablations_summaries.tex
\resizebox*{\linewidth}{!}{
\begin{tabular}{@{}clrrrrrrrrrrrrrrr@{}}
\toprule
\multirow{2}{*}{\textbf{Openness}} &
  \multicolumn{1}{c}{\multirow{2}{*}{\shortstack{\textbf{Summary Generation} \\ \textbf{Model}}}} &
  \multicolumn{5}{c}{\textbf{FEDI-Error}} &
  \multicolumn{5}{c}{\textbf{ABCEval}} &
  \multicolumn{5}{c}{\textbf{Soda-Eval}} \\ 
  \cmidrule(l){3-7} 
  \cmidrule(l){8-12}
  \cmidrule(l){13-17}
 &
  \multicolumn{1}{c}{} &
  \multicolumn{1}{c}{H-Score} &
  \multicolumn{1}{c}{Acc-K$^{\phantom{\dagger}}$\scriptsize\phantom{($\Uparrow$.00)}} &
  \multicolumn{1}{c}{Acc-U$^{\phantom{\dagger}}$\scriptsize\phantom{($\Uparrow$.00)}} &
  \multicolumn{1}{c}{ARI} &
  \multicolumn{1}{c}{NMI} &
  \multicolumn{1}{c}{H-Score} &
  \multicolumn{1}{c}{Acc-K$^{\phantom{\dagger}}$\scriptsize\phantom{($\Uparrow$.00)}} &
  \multicolumn{1}{c}{Acc-U$^{\phantom{\dagger}}$\scriptsize\phantom{($\Uparrow$.00)}} &
  \multicolumn{1}{c}{ARI} &
  \multicolumn{1}{c}{NMI} &
  \multicolumn{1}{c}{H-Score} &
  \multicolumn{1}{c}{Acc-K$^{\phantom{\dagger}}$\scriptsize\phantom{($\Uparrow$.00)}} &
  \multicolumn{1}{c}{Acc-U$^{\phantom{\dagger}}$\scriptsize\phantom{($\Uparrow$.00)}} &
  \multicolumn{1}{c}{ARI} &
  \multicolumn{1}{c}{NMI} \\ \midrule
\multirow{3}{*}{25\%}
                      & Llama-3.1 Instruct (8B) & 0.38 & 0.41$\phantom{^\mathbf{\dagger}}$\phantom{\scriptsize(\textcolor{dark_green}{$\Uparrow$.22})} & 0.34$\phantom{^\mathbf{\dagger}}$\phantom{\scriptsize(\textcolor{dark_green}{$\Uparrow$.23})} & 0.19 & 0.19 & 0.53 & 0.46$\phantom{^\mathbf{\dagger}}$\phantom{\scriptsize(\textcolor{dark_red}{$\Downarrow$.01})} & 0.68$\phantom{^\mathbf{\dagger}}$\phantom{\scriptsize(\textcolor{dark_green}{$\Uparrow$.43})} & 0.21 & 0.45 & 0.40 & 0.41 \phantom{\scriptsize(\textcolor{dark_green}{$\Uparrow$.08})} & 0.39$\phantom{^\mathbf{\dagger}}$\phantom{\scriptsize(\textcolor{dark_green}{$\Uparrow$.39})}& 0.15 & 0.17 \\ \cmidrule(l){2-17} 
                      & Phi-4-mini-instruct (3.84B)             & 0.37 & 0.39$\phantom{^\mathbf{\dagger}}$\scriptsize(\textcolor{dark_red}{$\Downarrow$.02}) & 0.32$\phantom{^\mathbf{\dagger}}$\scriptsize(\textcolor{dark_red}{$\Downarrow$.02}) & 0.17& 0.14& 0.51& 0.45$\phantom{^\mathbf{\dagger}}$\scriptsize(\textcolor{dark_red}{$\Downarrow$.01}) & 0.59$\phantom{^\mathbf{\dagger}}$\scriptsize(\textcolor{dark_red}{$\Uparrow$.09}) & 0.24 & 0.41 & 0.37 & 0.39$\phantom{^\mathbf{\dagger}}$\scriptsize(\textcolor{dark_red}{$\Downarrow$.02}) & 0.35$\phantom{^\mathbf{\dagger}}$\scriptsize(\textcolor{dark_red}{$\Downarrow$.04}) & 0.17& 0.19 \\ 
                      & DeepSeek-R1-Distill-Qwen (7B)                & 0.39 & 0.42$\phantom{^\mathbf{\dagger}}$\scriptsize(\textcolor{dark_green}{$\Uparrow$.01}) & 0.36$\phantom{^\mathbf{\dagger}}$\scriptsize(\textcolor{dark_green}{$\Uparrow$.02}) & 0.17 & 0.20& 0.60& 0.53$\phantom{^\mathbf{\dagger}}$\scriptsize(\textcolor{dark_green}{$\Uparrow$.07}) & 0.70$\phantom{^\mathbf{\dagger}}$\scriptsize(\textcolor{dark_green}{$\Uparrow$.02}) & 0.23& 0.40& 0.44& 0.46$\phantom{^\mathbf{\dagger}}$\scriptsize(\textcolor{dark_green}{$\Uparrow$.05}) & 0.42$\phantom{^\mathbf{\dagger}}$\scriptsize(\textcolor{dark_green}{$\Uparrow$.03})& 0.22 & 0.24\\ \midrule
\multirow{3}{*}{50\%}
                      & Llama-3.1 Instruct (8B) & 0.33 & 0.48$\phantom{^\mathbf{\dagger}}$\phantom{\scriptsize(\textcolor{dark_green}{$\Uparrow$.30})} & 0.22$\phantom{^\mathbf{\dagger}}$\phantom{\scriptsize(\textcolor{dark_green}{$\Uparrow$.05})} & 0.13 & 0.15 & 0.64 & 0.67$\phantom{^\mathbf{\dagger}}$\phantom{\scriptsize(\textcolor{dark_green}{$\Uparrow$.39})}& 0.62$\phantom{^\mathbf{\dagger}}$\phantom{\scriptsize(\textcolor{dark_green}{$\Uparrow$.20})}& 0.29 & 0.51 & 0.37 & 0.49$\phantom{^\mathbf{\dagger}}$\phantom{\scriptsize(\textcolor{dark_green}{$\Uparrow$.21})}& 0.30$\phantom{^\mathbf{\dagger}}$\phantom{\scriptsize(\textcolor{dark_green}{$\Uparrow$.11})}& 0.19 & 0.19 \\ \cmidrule(l){2-17} 
                      & Phi-4-mini-instruct (3.84B)             & 0.28 & 0.48$\phantom{^\mathbf{\dagger}}$\phantom{\scriptsize(\textcolor{dark_green}{$\Uparrow$.21})} & 0.20$\phantom{^\mathbf{\dagger}}$\scriptsize(\textcolor{dark_red}{$\Downarrow$.02}) & 0.10& 0.14& 0.57& 0.60$\phantom{^\mathbf{\dagger}}$\scriptsize(\textcolor{dark_red}{$\Downarrow$.07}) & 0.55$\phantom{^\mathbf{\dagger}}$\scriptsize(\textcolor{dark_red}{$\Downarrow$.07}) &  0.25&  0.45&  0.36& 0.55$\phantom{^\mathbf{\dagger}}$\scriptsize(\textcolor{dark_green}{$\Uparrow$.06}) & 0.27$\phantom{^\mathbf{\dagger}}$\scriptsize(\textcolor{dark_red}{$\Downarrow$.03}) & 0.22& 0.27\\ 
                      & DeepSeek-R1-Distill-Qwen (7B)                & 0.32& 0.52$\phantom{^\mathbf{\dagger}}$\scriptsize(\textcolor{dark_green}{$\Uparrow$.04}) & 0.23$\phantom{^\mathbf{\dagger}}$\scriptsize(\textcolor{dark_green}{$\Uparrow$.01}) & 0.11& 0.16& 0.64& 0.69$\phantom{^\mathbf{\dagger}}$\scriptsize(\textcolor{dark_green}{$\Uparrow$.02}) & 0.60$\phantom{^\mathbf{\dagger}}$\scriptsize(\textcolor{dark_red}{$\Downarrow$.02}) & 0.32 & 0.49& 0.37& 0.59$\phantom{^\mathbf{\dagger}}$\scriptsize(\textcolor{dark_green}{$\Uparrow$.10}) & 0.27$\phantom{^\mathbf{\dagger}}$\scriptsize(\textcolor{dark_green}{$\Downarrow$.03})& 0.24& 0.29 \\ \midrule
\multirow{3}{*}{75\%}
                      & Llama-3.1 Instruct (8B)     & 0.37 & 0.64$\phantom{^\mathbf{\dagger}}$\phantom{\scriptsize(\textcolor{dark_green}{$\Uparrow$.49})} & 0.26$\phantom{^\mathbf{\dagger}}$\phantom{\scriptsize(\textcolor{dark_green}{$\Uparrow$.09})} & 0.16 & 0.17 & 0.60 & 0.75$\phantom{^\mathbf{\dagger}}$\phantom{\scriptsize(\textcolor{dark_green}{$\Uparrow$.43})} & 0.50$\phantom{^\mathbf{\dagger}}$\phantom{\scriptsize(\textcolor{dark_green}{$\Uparrow$.01})} & 0.21 & 0.47 & 0.42 & 0.61$\phantom{^\mathbf{\dagger}}$\phantom{\scriptsize(\textcolor{dark_green}{$\Uparrow$.42})}& 0.32$\phantom{^\mathbf{\dagger}}$\phantom{\scriptsize(\textcolor{dark_green}{$\Uparrow$.01})}& 0.12 & 0.14 \\ \cmidrule(l){2-17} 
                      & Phi-4-mini-instruct (3.84B)             & 0.38& 0.63$\phantom{^\mathbf{\dagger}}$\scriptsize(\textcolor{dark_red}{$\Uparrow$.01}) & 0.27$\phantom{^\mathbf{\dagger}}$\scriptsize(\textcolor{dark_green}{$\Uparrow$.01}) & 0.16& 0.17& 0.54& 0.75$\phantom{^\mathbf{\dagger}}$\phantom{\scriptsize(\textcolor{dark_red}{$\Downarrow$.02})} & 0.42$\phantom{^\mathbf{\dagger}}$\scriptsize(\textcolor{dark_red}{$\Downarrow$.08}) & 0.23& 0.39& 0.34& 0.53$\phantom{^\mathbf{\dagger}}$\scriptsize(\textcolor{dark_red}{$\Downarrow$.08}) & 0.25$\phantom{^\mathbf{\dagger}}$\scriptsize(\textcolor{dark_red}{$\Downarrow$.07}) & 0.08& 0.12\\ 
                      & DeepSeek-R1-Distill-Qwen (7B)                & 0.44& 0.69$\phantom{^\mathbf{\dagger}}$\scriptsize(\textcolor{dark_green}{$\Uparrow$.05}) & 0.32$\phantom{^\mathbf{\dagger}}$\scriptsize(\textcolor{dark_green}{$\Uparrow$.06}) & 0.19& 0.25& 0.60& 0.77$\phantom{^\mathbf{\dagger}}$\scriptsize(\textcolor{dark_green}{$\Uparrow$.02}) & 0.49$\phantom{^\mathbf{\dagger}}$\scriptsize(\textcolor{dark_red}{$\Downarrow$.01}) & 0.30 & 0.52& 0.43& 0.60$\phantom{^\mathbf{\dagger}}$\scriptsize(\textcolor{dark_red}{$\Downarrow$.01}) & 0.34$\phantom{^\mathbf{\dagger}}$\scriptsize(\textcolor{dark_green}{$\Uparrow$.02}) & 0.12& 0.15 \\ \bottomrule
\end{tabular}
}



%% file: appendix/tables/summary_generation.tex
\resizebox*{\linewidth}{!}{
\begin{tabular}{@{}lrrrrrrrrr@{}}
\toprule
\multicolumn{1}{c}{\multirow{2}{*}{\textbf{Model}}} &
  \multicolumn{3}{c}{\textbf{FEDI-Error}} &
  \multicolumn{3}{c}{\textbf{ABCEval}} &
  \multicolumn{3}{c}{\textbf{Soda-Eval}} \\   \cmidrule(l){2-4} 
  \cmidrule(l){5-7}
  \cmidrule(l){8-10}
\multicolumn{1}{c}{} &
  \multicolumn{1}{l}{Faith.} &
  \multicolumn{1}{l}{Comp.} &
  \multicolumn{1}{l}{Conc.} &
  \multicolumn{1}{l}{Faith.} &
  \multicolumn{1}{l}{Comp.} &
  \multicolumn{1}{l}{Conc.} &
  \multicolumn{1}{l}{Faith.} &
  \multicolumn{1}{l}{Comp.} &
  \multicolumn{1}{l}{Conc.} \\ \midrule
Llama-3.1 & 0.67 & 0.63 & 0.59 & 0.51 & 0.59 & 0.53 & 0.62 & 0.66 & 0.54 \\
Phi-4  & 0.61 & 0.61 & 0.57 & 0.45 & 0.58 & 0.46 & 0.48 & 0.60 & 0.49 \\
\textbf{DeepSeek-R1} & \textbf{0.68} & \textbf{0.72} & \textbf{0.61} & \textbf{0.59} & \textbf{0.63} & \textbf{0.57} & \textbf{0.72} & \textbf{0.70} & \textbf{0.64} \\ \bottomrule
\end{tabular}
}



%% file: appendix/tables/syncid_loop_ablations.tex
\resizebox*{\linewidth}{!}{
\begin{tabular}{@{}clrrrrrrrrrrrrrrr@{}}
\toprule
\multirow{2}{*}{\textbf{Openness}} &
  \multicolumn{1}{c}{\multirow{2}{*}{\textbf{Method}}} &
  \multicolumn{5}{c}{\textbf{FEDI-Error}} &
  \multicolumn{5}{c}{\textbf{ABCEval}} &
  \multicolumn{5}{c}{\textbf{Soda-Eval}} \\ \cmidrule(l){3-17} 
 &
  \multicolumn{1}{c}{} &
  \multicolumn{1}{c}{H-Score} &
  \multicolumn{1}{c}{Acc-K$\phantom{\ _{050\%}}$} &
  \multicolumn{1}{c}{Acc-U$\phantom{\ _{050\%}}$} &
  \multicolumn{1}{c}{ARI} &
  \multicolumn{1}{c}{NMI} &
  \multicolumn{1}{c}{H-Score} &
  \multicolumn{1}{c}{Acc-K$\phantom{\ _{050\%}}$} &
  \multicolumn{1}{c}{Acc-U$\phantom{\ _{050\%}}$} &
  \multicolumn{1}{c}{ARI} &
  \multicolumn{1}{c}{NMI} &
  \multicolumn{1}{c}{H-Score} &
  \multicolumn{1}{c}{Acc-K$\phantom{\ _{050\%}}$} &
  \multicolumn{1}{c}{Acc-U$\phantom{\ _{050\%}}$} &
  \multicolumn{1}{c}{ARI} &
  \multicolumn{1}{c}{NMI} \\ \midrule
\multirow{6}{*}{25\%} & SynCID & 0.27 & 0.40 \scriptsize\phantom{($\Uparrow$.00)} & 0.20 \scriptsize\phantom{($\Uparrow$.00)} & 0.06 & 0.11 & 0.53 & 0.45 \scriptsize\phantom{($\Uparrow$.00)} & 0.68 \scriptsize\phantom{($\Uparrow$.00)} & 0.03 & 0.41 & 0.31 & 0.38 \scriptsize\phantom{($\Uparrow$.00)}& 0.26 \scriptsize\phantom{($\Uparrow$.00)} & 0.11 & 0.14 \\ 
                      & \hspace{0.5em} w/o Stage 2 & 0.27 & 0.40 \scriptsize\phantom{($\Uparrow$.00)} & 0.20 \scriptsize\phantom{($\Uparrow$.00)} & 0.06 & 0.11 & 0.50 & 0.44 \scriptsize(\textcolor{dark_red}{$\Downarrow$.01}) & 0.64 \scriptsize(\textcolor{dark_red}{$\Downarrow$.04}) & 0.04 & 0.42 & 0.31 & 0.35 \scriptsize(\textcolor{dark_red}{$\Downarrow$.03}) & 0.27 \scriptsize(\textcolor{dark_green}{$\Uparrow$.01}) & 0.10 & 0.14 \\ \cmidrule(l){2-17} 
                      & LOOP (LIS) & 0.26 & 0.37 \scriptsize\phantom{($\Uparrow$.00)} & 0.19 \scriptsize\phantom{($\Uparrow$.00)} & 0.09 & 0.10 & 0.51 & 0.43 \scriptsize\phantom{($\Uparrow$.00)} & 0.63 \scriptsize\phantom{($\Uparrow$.00)} & 0.01 & 0.37 & 0.33 & 0.36 \scriptsize\phantom{($\Uparrow$.00)} & 0.31 \scriptsize\phantom{($\Uparrow$.00)} & 0.07 & 0.13 \\ 
                      & \hspace{0.5em} w/o Stage 2 & 0.25 & 0.34 \scriptsize(\textcolor{dark_red}{$\Downarrow$.03}) & 0.20 \scriptsize(\textcolor{dark_green}{$\Uparrow$.01}) & 0.06 & 0.08 & 0.46 & 0.38 \scriptsize(\textcolor{dark_red}{$\Downarrow$.05}) & 0.60 \scriptsize(\textcolor{dark_red}{$\Downarrow$.03}) & 0.01 & 0.38 & 0.28 & 0.35 \scriptsize(\textcolor{dark_red}{$\Downarrow$.01}) & 0.24 \scriptsize(\textcolor{dark_red}{$\Downarrow$.07}) & 0.05 & 0.11 \\ 
                      & LOOP (LBSR) & 0.28 & 0.36 \scriptsize(\textcolor{dark_red}{$\Downarrow$.01}) & 0.23 \scriptsize(\textcolor{dark_green}{$\Uparrow$.04}) & 0.11 & 0.10 & 0.61 & 0.55 \scriptsize(\textcolor{dark_green}{$\Uparrow$.12}) &  0.68 \scriptsize(\textcolor{dark_green}{$\Uparrow$.05}) & 0.06 & 0.43 & 0.34 & 0.38 \scriptsize(\textcolor{dark_green}{$\Uparrow$.02}) & 0.30 \scriptsize(\textcolor{dark_red}{$\Downarrow$.01}) & 0.09 & 0.14 \\ \cmidrule(l){2-17}
                      & SEEED & 0.38 & 0.41 \scriptsize\phantom{($\Uparrow$.00)} & 0.34 \scriptsize\phantom{($\Uparrow$.00)} & 0.19 & 0.19 & 0.53 & 0.46 \scriptsize\phantom{($\Uparrow$.00)} & 0.68 \scriptsize\phantom{($\Uparrow$.00)} & 0.21 & 0.45 & 0.40 & 0.41 \scriptsize\phantom{($\Uparrow$.00)} & 0.39 \scriptsize\phantom{($\Uparrow$.00)} & 0.15 & 0.17 \\ \midrule
\multirow{6}{*}{50\%} & SynCID & 0.26 & 0.34 \scriptsize\phantom{($\Uparrow$.00)} & 0.21 \scriptsize\phantom{($\Uparrow$.00)} & 0.04 & 0.09 & 0.59 & 0.55 \scriptsize\phantom{($\Uparrow$.00)} & 0.64 \scriptsize\phantom{($\Uparrow$.00)} & 0.11 & 0.47 & 0.27 & 0.40 \scriptsize\phantom{($\Uparrow$.00)}& 0.21 \scriptsize\phantom{($\Uparrow$.00)} & 0.09 & 0.11 \\ 
                      & \hspace{0.5em} w/o Stage 2     & 0.26 & 0.28 \scriptsize(\textcolor{dark_red}{$\Downarrow$.06}) & 0.24 \scriptsize(\textcolor{dark_green}{$\Uparrow$.03})& 0.03 & 0.07 & 0.53 & 0.46 \scriptsize(\textcolor{dark_red}{$\Downarrow$.09}) & 0.65 \scriptsize(\textcolor{dark_green}{$\Downarrow$.01}) & 0.10 & 0.46 & 0.26 & 0.40 \scriptsize\phantom{(\textcolor{dark_red}{$\Downarrow$.03})} & 0.19 \scriptsize(\textcolor{dark_red}{$\Downarrow$.02}) & 0.08 & 0.11 \\ \cmidrule(l){2-17} 
                      & LOOP (LIS) & 0.22 & 0.39 \scriptsize\phantom{($\Uparrow$.00)} & 0.16 \scriptsize\phantom{($\Uparrow$.00)} & 0.07 & 0.07 & 0.45 & 0.48 \scriptsize\phantom{($\Uparrow$.00)} & 0.43 \scriptsize\phantom{($\Uparrow$.00)} & 0.03 & 0.41 & 0.24 & 0.55 \scriptsize\phantom{($\Uparrow$.00)} & 0.16 \scriptsize\phantom{($\Uparrow$.00)} & 0.11 & 0.16 \\ 
                      & \hspace{0.5em} w/o Stage 2 & 0.21 & 0.36 \scriptsize(\textcolor{dark_red}{$\Downarrow$.03}) & 0.15 \scriptsize(\textcolor{dark_red}{$\Downarrow$.01}) & 0.04 & 0.07 & 0.37 & 0.42 \scriptsize(\textcolor{dark_red}{$\Downarrow$.07}) & 0.36 \scriptsize(\textcolor{dark_red}{$\Downarrow$.07}) & 0.03 & 0.40 & 0.25 & 0.49 \scriptsize(\textcolor{dark_red}{$\Downarrow$.06}) & 0.17 \scriptsize(\textcolor{dark_green}{$\Uparrow$.01}) & 0.09 & 0.15 \\ 
                      & LOOP (LBSR) & 0.25 & 0.40 \scriptsize(\textcolor{dark_green}{$\Uparrow$.01}) & 0.18 \scriptsize(\textcolor{dark_green}{$\Uparrow$.02}) & 0.06 & 0.07 & 0.46 & 0.58 \scriptsize(\textcolor{dark_green}{$\Uparrow$.10}) & 0.41 \scriptsize(\textcolor{dark_red}{$\Downarrow$.02}) & 0.08 & 0.46 & 0.25 & 0.58 \scriptsize(\textcolor{dark_green}{$\Uparrow$.03}) & 0.16 \scriptsize\phantom{($\Uparrow$.00)} & 0.13 & 0.17 \\ \cmidrule(l){2-17}
                      & SEEED & 0.33 & 0.48 \scriptsize\phantom{($\Uparrow$.00)} & 0.22 \scriptsize\phantom{($\Uparrow$.00)} & 0.13 & 0.15 & 0.64 & 0.67 \scriptsize\phantom{($\Uparrow$.00)} & 0.62 \scriptsize\phantom{($\Uparrow$.00)} & 0.29 & 0.51 & 0.37 & 0.49 \scriptsize\phantom{($\Uparrow$.00)} & 0.30 \scriptsize\phantom{($\Uparrow$.00)} & 0.19 & 0.19 \\ \midrule
\multirow{6}{*}{75\%} & SynCID & 0.23 & 0.36 \scriptsize\phantom{($\Uparrow$.00)} & 0.17 \scriptsize\phantom{($\Uparrow$.00)} & 0.06 & 0.01 & 0.54 & 0.59 \scriptsize\phantom{($\Uparrow$.00)} & 0.50 \scriptsize\phantom{($\Uparrow$.00)} & 0.07 & 0.44 & 0.25 & 0.22 \scriptsize\phantom{($\Uparrow$.00)}& 0.28 \scriptsize\phantom{($\Uparrow$.00)} & 0.02 & 0.06 \\ 
                      & \hspace{0.5em} w/o Stage 2     & 0.22 & 0.35 \scriptsize(\textcolor{dark_red}{$\Downarrow$.01}) & 0.16 \scriptsize(\textcolor{dark_red}{$\Uparrow$.01}) & 0.01 & 0.06 & 0.54 & 0.58 \scriptsize(\textcolor{dark_red}{$\Downarrow$.01}) & 0.51 \scriptsize(\textcolor{dark_green}{$\Uparrow$.01})& 0.09 & 0.45 & 0.24 & 0.27 \scriptsize(\textcolor{dark_green}{$\Uparrow$.05}) & 0.15 \scriptsize(\textcolor{dark_red}{$\Downarrow$.13}) & 0.02 & 0.04\\ \cmidrule(l){2-17} 
                      & LOOP (LIS) & 0.25 & 0.43 \scriptsize\phantom{($\Uparrow$.00)} & 0.18 \scriptsize\phantom{($\Uparrow$.00)} & 0.05 & 0.01 & 0.48 & 0.69 \scriptsize\phantom{($\Uparrow$.00)} & 0.37 \scriptsize\phantom{($\Uparrow$.00)} & 0.07 & 0.44 & 0.22 & 0.31 \scriptsize\phantom{($\Uparrow$.00)} & 0.17 \scriptsize\phantom{($\Uparrow$.00)} & 0.07 & 0.08 \\ 
                      & \hspace{0.5em} w/o Stage 2 & 0.21 & 0.39 \scriptsize(\textcolor{dark_red}{$\Downarrow$.04}) & 0.14 \scriptsize(\textcolor{dark_red}{$\Downarrow$.04}) & 0.01 & 0.05 & 0.43 & 0.64 \scriptsize(\textcolor{dark_red}{$\Downarrow$.05}) & 0.34 \scriptsize(\textcolor{dark_red}{$\Downarrow$.03}) & 0.03 & 0.40 & 0.22 & 0.29 \scriptsize(\textcolor{dark_red}{$\Downarrow$.02}) & 0.18 \scriptsize(\textcolor{dark_green}{$\Uparrow$.01}) & 0.07 & 0.09 \\
                      & LOOP (LBSR) & 0.25 & 0.44 \scriptsize(\textcolor{dark_green}{$\Uparrow$.01}) & 0.17 \scriptsize(\textcolor{dark_red}{$\Downarrow$.01}) & 0.01 & 0.05 & 0.51 & 0.71 \scriptsize(\textcolor{dark_green}{$\Uparrow$.02}) &  0.40 \scriptsize(\textcolor{dark_green}{$\Uparrow$.03}) & 0.08 & 0.45 & 0.26 & 0.43 \scriptsize(\textcolor{dark_green}{$\Uparrow$.12}) & 0.19 \scriptsize(\textcolor{dark_green}{$\Uparrow$.02}) & 0.11 & 0.08 \\ \cmidrule(l){2-17}
                      & SEEED & 0.37 & 0.64 \scriptsize\phantom{($\Uparrow$.00)} & 0.26 \scriptsize\phantom{($\Uparrow$.00)} & 0.16 & 0.17 & 0.60 & 0.75 \scriptsize\phantom{($\Uparrow$.00)} & 0.50 \scriptsize\phantom{($\Uparrow$.00)} & 0.21 & 0.47 & 0.42 & 0.61 \scriptsize\phantom{($\Uparrow$.00)} & 0.32 \scriptsize\phantom{($\Uparrow$.00)} & 0.12 & 0.14 \\ \bottomrule
\end{tabular}
}



%% file: appendix/tables/fedi_cluster_interpretation_1.tex
\resizebox{0.9\linewidth}{!}{%
    \begin{tabular}{p{0.40\linewidth}p{0.40\linewidth}c@{}} 
        \toprule        
        \multicolumn{1}{c}{\textbf{Ground Truth}} &
        \multicolumn{1}{c}{\textbf{Generated}} &
        \multicolumn{1}{c}{\textbf{Acc-U}} \\
        \midrule  
                {\footnotesize \textbf{Ignore Question} When the system fails to address the user's question, this is called an ignore question error. Instead of providing a relevant response or clarification, the system disregards ...} &
        {\footnotesize \textbf{Unaddressed Request} The system neglects the user's question, failing to provide a relevant response. This can lead to frustration and ultimately interrupt the conversation.} &
        {\footnotesize 0.21} \\
        \midrule
        {\footnotesize \textbf{Lack of Sociality} When the system's responses overlook social conventions and fail to include basic greetings or exhibit toxic and disrespectful behavior or language, this is referred to as a lack of sociality error. ...} &
        {\footnotesize \textbf{Insensitive Interaction} This error occurs when a system's responses disregard social norms, exhibit impoliteness, or employ toxic and condescending language. ...} &
        {\footnotesize 0.24} \\ 
        \midrule        
        {\footnotesize \textbf{Factually Incorrect} If the response provided by the system contains information that is factually wrong or inaccurate, this is referred to as a factually incorrect error. ...} &
        {\footnotesize \textbf{Misinformation} When the system provides incorrect information, this is called misinformation. This can happen when the system's world knowledge is outdated, incomplete, or simply wrong. ...} &
        {\footnotesize 0.20} \\
        \midrule
        {\footnotesize \textbf{Ignore Request} When the system fails to take action on a user's request, this is called an ignore request error. This error can occur due to various reasons, such as misinterpretation of the request, technical ...} &
        {\footnotesize \textbf{Disregarded Request} The system does not directly address the user's request. This can happen due to misunderstandings or system issues, leading to user frustration and communication breakdown.} &
        {\footnotesize 0.33} \\
        \midrule    
        {\footnotesize \textbf{Attribute Error} When the system fails to correctly extract or understand the necessary slots or attributes from the user's utterance, this is called an attribute ...} &
        {\footnotesize \textbf{Attribute Error} When the system fails to accurately extract or understand necessary information from a user utterance that is necessary for task completion.} &
        {\footnotesize 0.27} \\
        \bottomrule
    \end{tabular}
}

%% file: appendix/tables/fedi_cluster_interpretation_2.tex
\resizebox{0.9\linewidth}{!}{%
    \begin{tabular}{p{0.40\linewidth}p{0.40\linewidth}c@{}} 
        \toprule        
        \multicolumn{1}{c}{\textbf{Ground Truth}} &
        \multicolumn{1}{c}{\textbf{Generated}} &
        \multicolumn{1}{c}{\textbf{Acc-U}} \\
                \midrule
        {\footnotesize \textbf{Topic Transition Error} If the system's response abruptly shifts to a different or previously discussed topic without a logical connection or adequate context, this is called a topic transition error. This error disrupts the flow and coherence of the conversation, causing confusion and frustration for the human ...} &
        {\footnotesize \textbf{Abrupt Topic Shift} An abrupt topic shift error occurs when a system's response suddenly changes to a new topic without a clear connection to the current context. This can disrupt the conversation flow and cause confusion, leading to frustration and decreased trust in the system's responses.} &
        {\footnotesize 0.28} \\  
        \midrule          
        {\footnotesize \textbf{Conversationality} Bad conversationality occurs when the system fails to maintain a coherent and natural conversation flow, e.g., the system repeats its previous responses or contradicts itself without recognizing or asking for new or missing information. ... } &
        {\footnotesize \textbf{Inconsistency} When the system's responses lack coherence, often repeating itself or contradicting previous statements without seeking or acknowledging new or missing information. This leads to poor communication, damaging user trust and confidence. ...} &
        {\footnotesize 0.40} \\
        
        \bottomrule
    \end{tabular}
}

%% file: appendix/tables/fedi_cluster_interpretation_3.tex
\resizebox{0.9\linewidth}{!}{%
    \begin{tabular}{p{0.40\linewidth}p{0.40\linewidth}c@{}} 
        \toprule        
        \multicolumn{1}{c}{\textbf{Ground Truth}} &
        \multicolumn{1}{c}{\textbf{Generated}} &
        \multicolumn{1}{c}{\textbf{Acc-U}} \\
        \midrule         
        {\footnotesize \textbf{Unclear Intention} When the system fails to accurately comprehend and address the user's intended objective, this is referred to as an unclear intention error. This error often arises due to ambiguous or incomplete user inputs, conflicting context, or limitations ...} &
        {\footnotesize \textbf{Misaligned Goal} A misaligned goal occurs when the system misinterprets the user's objective, often due to unclear or conflicting user input. This error can stem from the user's input being ambiguous, incomplete, or inconsistent with the context. ...} &
        {\footnotesize 0.33} \\ \midrule
        {\footnotesize \textbf{Ignore Expectation} When the system's fails to meet the user's expectation, this is called an ignore expectation error. In this error type, the system either overlooks or disregards important information provided by the user, resulting in an incomplete response. ...} &
        {\footnotesize \textbf{Misaligned Response} A system response that fails to accurately understand or address the user's needs. } &
        {\footnotesize 0.31} \\ \bottomrule
    \end{tabular}
}

%% file: appendix/tables/abceval_cluster_interpretation_1.tex
\resizebox{0.9\linewidth}{!}{%
    \begin{tabular}{p{0.40\linewidth}p{0.40\linewidth}c@{}} 
        \toprule        
        \multicolumn{1}{c}{\textbf{Ground Truth}} &
        \multicolumn{1}{c}{\textbf{Generated}} &
        \multicolumn{1}{c}{\textbf{Acc-U}} \\
        \midrule        
        {\footnotesize \textbf{Uninterpretable} A response is uninterpretable if it is difficult to understand the intended meaning of part or all of the response in the context of the dialogue.} &
        {\footnotesize \textbf{Ambiguous} A response is ambiguous if parts of it are unclear in the dialogue context.} &
        {\footnotesize 1.0} \\
        \midrule        
        {\footnotesize \textbf{Ignore} Responses that are completely off-topic, fail to address the asked question, or are otherwise completely inappropriate in the context are considered to be ignoring the other speaker.} &
        {\footnotesize \textbf{Off-Topic Response} The response deviates from the topic, fails to answer the posed question, or is contextually inappropriate, indicating a disregard for the other speaker.} &
        {\footnotesize 0.61} \\
        \midrule        
        {\footnotesize \textbf{Commonsense Contradiction} To identify contradictions of commonsense, judge whether a vast majority of people would agree that the response doesn’t make sense because the response: ...} &
        {\footnotesize \textbf{Inconsistent Reasoning} A response that contains significant logical flaws or contradictions, goes against the general understanding of most people, or makes assumptions without a solid basis.} &
        {\footnotesize 0.63} \\   \midrule
        {\footnotesize \textbf{Incorrect Fact} Incorrect facts occur when the response includes information that is either: (1) false, (2) unproven, (3) highly controversial, (4) highly implausible, (5) clearly misleading. If an organization, person, place, etc. ...} &
        {\footnotesize \textbf{Misinformation} Misinformation occurs when a turn contains information that is not verifiable. A turn could be considered misinformed if it inaccurately represents historical facts, organizations, persons, or places.} &
        {\footnotesize 0.50} \\
        \midrule
        {\footnotesize \textbf{Self Contradiction} Self contradictions occur when the system says something that is a contradiction of what they have said previously or it is extremely implausible based on ...} &
        {\footnotesize \textbf{Self Contradiction} Speaker 2 provides information that contradicts previous statements or is implausible given the context. This can happen within a single turn or across multiple turns.} &
        {\footnotesize 0.50} \\     \midrule
            {\footnotesize \textbf{Partner Contradiction} Partner contradictions occur when the system: (1) shares an assumption about the user that ...} &
        {\footnotesize \textbf{Misunderstanding} When the system makes assumptions about the user that are impossible to infer from the conversation, asks repetitive questions or exhibits memory lapses about previsous user statements.} &
        {\footnotesize 1.0} \\  
        \bottomrule
    \end{tabular}
}

%% file: appendix/tables/abceval_cluster_interpretation_3.tex
\resizebox{0.9\linewidth}{!}{%
    \begin{tabular}{p{0.40\linewidth}p{0.40\linewidth}c@{}} 
        \toprule        
        \multicolumn{1}{c}{\textbf{Ground Truth}} &
        \multicolumn{1}{c}{\textbf{Generated}} &
        \multicolumn{1}{c}{\textbf{Acc-U}} \\
\midrule
        {\footnotesize \textbf{Redundant} A response is repetitive if: (1) it repeats something from earlier in the dialogue, (2) it includes asking a question whose answer has been already shared. If any part of the response is repetitive, then it should be labeled as repetitive. Note that sometimes repetition is useful, such as for emphasis, acknowledgement, clarification, or elaboration, and ..} &
        {\footnotesize \textbf{Unnecessary Repeat} A response is redundant if it repeats information already shared or asks a question with a known answer. This can occur in various forms, such as direct repetition, rephrased questions, or unnecessary elaboration. However, repetition can be justified in cases where it serves a purpose like emphasis, acknowledgment, or additional explanation.} &
        {\footnotesize 0.50} \\ \midrule
        {\footnotesize \textbf{Lack of Empathy} A response displays a lack of empathy when: (1) it indicates a misunderstanding of how the user feels based on what Speaker 1 just said, (2) the tone, emotion, or sentiment of the response is clearly inappropriate for what the user just said, (3) ...} &
        {\footnotesize \textbf{Insensitive Response} Insensitive responses occur when a speaker fails to acknowledge the user's emotions, often due to a misinterpretation of their feelings or an inappropriate tone. This can be seen in responses that seem dismissive or unemotional.} &
        {\footnotesize 0.43} \\
        \midrule
        {\footnotesize \textbf{Irrelevant} If a response fails to continue the current discussion or jumps to a new and off-topic discussion, it is considered to be irrelevant. Responses that are irrelevant feel abrupt and interrupt the discussion, ...} &
        {\footnotesize \textbf{Disconnected Response} A response is considered disconnected if it fails to build upon the previous turn, instead introducing a new topic or question. This type of response can disrupt the conversation flow.} &
        {\footnotesize 0.40} \\
        \bottomrule
    \end{tabular}
}

%% file: appendix/tables/soda_eval_cluster_interpretation_1.tex
\resizebox{0.9\linewidth}{!}{%
    \begin{tabular}{p{0.40\linewidth}p{0.40\linewidth}c@{}} 
        \toprule        
        \multicolumn{1}{c}{\textbf{Ground Truth}} &
        \multicolumn{1}{c}{\textbf{Generated}} &
        \multicolumn{1}{c}{\textbf{Acc-U}} \\
        \midrule         
        {\footnotesize \textbf{Coherence} Contradicts or ignores prior information in the dialogue.} &
        {\footnotesize \textbf{Inconsistency} Fails to maintain a logical connection with previous statements.} &
        {\footnotesize 0.18} \\
        \midrule  
                     
        {\footnotesize \textbf{Antisocial} Contains unsafe or inappropriate behaviour.} &
        {\footnotesize \textbf{Disrespectful} Characterized by the use of offensive language, derogatory terms, and aggressive tone, which can cause emotional distress.} &
        {\footnotesize 0.33} \\
        \midrule

                {\footnotesize \textbf{Fluency} Contains typos or other grammatical errors.} &
        {\footnotesize \textbf{Clarity} The response from speaker 2 contains spelling/grammar errors.} &
        {\footnotesize 0.30} \\ \midrule
                {\footnotesize \textbf{Commonsense} Lacks common knowledge and logic.} &
        {\footnotesize \textbf{Missing World Knowledge} Fails to demonstrate basic understanding of the world. In the context of a set of dyadic dialogues, this error type might manifest as conversations where one participant expects the other to possess knowledge or behave in a way that is not grounded in reality.} &
        {\footnotesize 0.14} \\ \midrule
        {\footnotesize \textbf{Gender Pronoun} Goes against normative pronoun.} &
        {\footnotesize \textbf{Gender Pronoun Mismatch} The use of pronouns that do not consistently align with the gender identity of the individuals being referred to result in a mismatch between the pronouns used and the gender norms expected in the dialogue.} &
        {\footnotesize 0.29} \\
        \midrule
        {\footnotesize \textbf{Non Textual} Includes narrative elements or references unexpected inside a turn of a dyadic interaction.} &
        {\footnotesize \textbf{Narrative Elements} The responses contain narrative elements or references that are not coherent within a round of dyadic interaction and may disrupt the expected flow of the dialogue.} &
        {\footnotesize 0.29} \\ \midrule
                          
        {\footnotesize \textbf{Engagement} Lacks a behaviour or emotion expected from the situation.} &
        {\footnotesize \textbf{Emotional Dissonance} The response lacks a behaviour or emotion that is typically associated with the situation, leading to an incongruous tone or atmosphere.} &
        {\footnotesize 0.39} \\       
        
        \bottomrule
    \end{tabular}
}

%% file: appendix/tables/soda_eval_cluster_interpretation_3.tex
\resizebox{0.9\linewidth}{!}{%
    \begin{tabular}{p{0.40\linewidth}p{0.40\linewidth}c@{}} 
        \toprule        
        \multicolumn{1}{c}{\textbf{Ground Truth}} &
        \multicolumn{1}{c}{\textbf{Generated}} &
        \multicolumn{1}{c}{\textbf{Acc-U}} \\
        \midrule
        {\footnotesize \textbf{Repetition} Repeats prior information in the dialogue.} &
        {\footnotesize \textbf{Redundancy} This error occurs when a speaker unnecessarily repeats information that has already been stated in the dialogue, failing to provide new or relevant information, or simply rephrasing what has already been said.} &
        {\footnotesize 0.15} \\
          \midrule
        {\footnotesize \textbf{Assumption} Infers information not available in the dialogue context.} &
        {\footnotesize \textbf{Misattribution} A response that incorrectly assigns information or characteristics to a dialogue participant, entity, or context that is not explicitly stated or implied within the dialogue.} &
        {\footnotesize 0.24} \\      

        \bottomrule
    \end{tabular}
}

%% file: appendix/tables/intent_detection.tex
\resizebox*{\linewidth}{!}{
\begin{tabular}{@{}clrrrrrrrrrrrrrrr@{}}
\toprule
\multirow{2}{*}{\textbf{Openness}} &
  \multicolumn{1}{c}{\multirow{2}{*}{\textbf{Method}}} &
  \multicolumn{5}{c}{\textbf{CLINC}} &
  \multicolumn{5}{c}{\textbf{BANKING}} &
  \multicolumn{5}{c}{\textbf{StackOverflow}} \\ \cmidrule(l){3-17} 
 &
  \multicolumn{1}{c}{} &
  \multicolumn{1}{c}{H-Score} &
  \multicolumn{1}{c}{Acc-K$^{\phantom{\dagger}}$\scriptsize\phantom{($\Uparrow$.00)}} &
  \multicolumn{1}{c}{Acc-U$^{\phantom{\dagger}}$\scriptsize\phantom{($\Uparrow$.00)}} &
  \multicolumn{1}{c}{ARI} &
  \multicolumn{1}{c}{NMI} &
  \multicolumn{1}{c}{H-Score} &
  \multicolumn{1}{c}{Acc-K$^{\phantom{\dagger}}$\scriptsize\phantom{($\Uparrow$.00)}} &
  \multicolumn{1}{c}{Acc-U$^{\phantom{\dagger}}$\scriptsize\phantom{($\Uparrow$.00)}} &
  \multicolumn{1}{c}{ARI} &
  \multicolumn{1}{c}{NMI} &
  \multicolumn{1}{c}{H-Score} &
  \multicolumn{1}{c}{Acc-K$^{\phantom{\dagger}}$\scriptsize\phantom{($\Uparrow$.00)}} &
  \multicolumn{1}{c}{Acc-U$^{\phantom{\dagger}}$\scriptsize\phantom{($\Uparrow$.00)}} &
  \multicolumn{1}{c}{ARI} &
  \multicolumn{1}{c}{NMI} \\ \midrule
\multirow{4}{*}{25\%} & KNN-Contrastive     & 0.67 & 0.91$^{\phantom{\dagger}}$\scriptsize\phantom{($\Uparrow$.00)}& 0.53$^{\phantom{\dagger}}$\scriptsize\phantom{($\Uparrow$.00)}& 0.75 & 0.91 & 0.50 & 0.90$^{\phantom{\dagger}}$\scriptsize\phantom{($\Uparrow$.00)}& 0.34$^{\phantom{\dagger}}$\scriptsize\phantom{($\Uparrow$.00)}& 0.68 & 0.87 & 0.45 & 0.84$^{\phantom{\dagger}}$\scriptsize\phantom{($\Uparrow$.00)}& 0.31$^{\phantom{\dagger}}$\scriptsize\phantom{($\Uparrow$.00)}& 0.56 & 0.73 \\ 
                      & SynCID              & 0.80 & \textbf{0.95}$^{\phantom{\dagger}}$\scriptsize{(\textcolor{dark_green}{$\Uparrow$.04})} & 0.69$^{\phantom{\dagger}}$\scriptsize{(\textcolor{dark_green}{$\Uparrow$.16})} & 0.83 & 0.94 & 0.64 & 0.87$^{\phantom{\dagger}}$\scriptsize{(\textcolor{dark_red}{$\Downarrow$.03})} & 0.50$^{\phantom{\dagger}}$\scriptsize{(\textcolor{dark_green}{$\Uparrow$.16})}& 0.70 & 0.89 & 0.72 & 0.86$^{\phantom{\dagger}}$\scriptsize{(\textcolor{dark_green}{$\Uparrow$.02})}& 0.62$^{\phantom{\dagger}}$\scriptsize{(\textcolor{dark_green}{$\Uparrow$.31})}& 0.66 & 0.78 \\ 
                      & LOOP                & \textbf{0.85} & 0.93$^{\phantom{\dagger}}$\scriptsize{(\textcolor{dark_green}{$\Uparrow$.02})}& \textbf{0.78}$^{\phantom{\dagger}}$\scriptsize{(\textcolor{dark_green}{$\Uparrow$.25})}& \textbf{0.85} & \textbf{0.95} & 0.63 & 0.90$^{\phantom{\dagger}}$\scriptsize\phantom{{(\textcolor{dark_green}{$\Uparrow$.16})}}& 0.48$^{\phantom{\dagger}}$\scriptsize{(\textcolor{dark_green}{$\Uparrow$.14})} & 0.73 & \textbf{0.90} & 0.73 & 0.89$^{\phantom{\dagger}}$\scriptsize{(\textcolor{dark_green}{$\Uparrow$.05})}& 0.62$^{\phantom{\dagger}}$\scriptsize{(\textcolor{dark_green}{$\Uparrow$.31})} & 0.73 & 0.82 \\ \cmidrule(l){2-17}       
                      & \textbf{\textit{SEEED}}       & 0.82 & 0.93$^{\phantom{\dagger}}$\scriptsize{(\textcolor{dark_green}{$\Uparrow$.02})} & 0.74$^{\phantom{\dagger}}$\scriptsize{(\textcolor{dark_green}{$\Uparrow$.21})}& 0.79 & 0.93 & \textbf{0.79} & \textbf{0.92}$^{\phantom{\dagger}}$\scriptsize{(\textcolor{dark_green}{$\Uparrow$.02})}& \textbf{0.70}$^\mathbf{\dagger}$\scriptsize{(\textcolor{dark_green}{$\Uparrow$.36})}& \textbf{0.77} & \textbf{0.90} & \textbf{0.87} & \textbf{0.90}$^{\phantom{\mathbf{\dagger}}}$\scriptsize{(\textcolor{dark_green}{$\Uparrow$.06})}& \textbf{0.84}$^\mathbf{\dagger}$\scriptsize{(\textcolor{dark_green}{$\Uparrow$.53})}& \textbf{0.77} & \textbf{0.83} \\ \midrule
\multirow{4}{*}{50\%} & KNN-Contrastive     & 0.62 & 0.87$^{\phantom{\dagger}}$\scriptsize\phantom{($\Uparrow$.00)}& 0.48$^{\phantom{\dagger}}$\scriptsize\phantom{($\Uparrow$.00)}& 0.60 & 0.86 & 0.58 & 0.80$^{\phantom{\dagger}}$\scriptsize\phantom{($\Uparrow$.00)}& 0.45$^{\phantom{\dagger}}$\scriptsize\phantom{($\Uparrow$.00)}& 0.53 & 0.81 & 0.65 & 0.82$^{\phantom{\dagger}}$\scriptsize\phantom{($\Uparrow$.00)}& 0.54$^{\phantom{\dagger}}$\scriptsize\phantom{($\Uparrow$.00)}& 0.51 & 0.67 \\ 
                      & SynCID              & 0.77 & \textbf{0.95}$^{\phantom{\dagger}}$\scriptsize{(\textcolor{dark_green}{$\Uparrow$.08})} & 0.64$^{\phantom{\dagger}}$\scriptsize{(\textcolor{dark_green}{$\Uparrow$.16})}& 0.71 & 0.90 & 0.66 & 0.85$^{\phantom{\dagger}}$\scriptsize{(\textcolor{dark_green}{$\Uparrow$.05})}& 0.54$^{\phantom{\dagger}}$\scriptsize{(\textcolor{dark_green}{$\Uparrow$.09})}& 0.60 & 0.84 & 0.72 & 0.76$^{\phantom{\dagger}}$\scriptsize{(\textcolor{dark_red}{$\Downarrow$.06})}& 0.69$^{\phantom{\dagger}}$\scriptsize{(\textcolor{dark_green}{$\Uparrow$.15})} & 0.52 & 0.71 \\ 
                      & LOOP                & 0.80 & \textbf{0.95}$^{\phantom{\dagger}}$\scriptsize{(\textcolor{dark_green}{$\Uparrow$.08})}& 0.69$^{\phantom{\dagger}}$\scriptsize{(\textcolor{dark_green}{$\Uparrow$.21})}& \textbf{0.75} & \textbf{0.92} & 0.63 & 0.90$^{\phantom{\dagger}}$\scriptsize{(\textcolor{dark_green}{$\Uparrow$.10})}& 0.48$^{\phantom{\dagger}}$\scriptsize{(\textcolor{dark_green}{$\Uparrow$.03})}& 0.63 & 0.86 & 0.80 & \textbf{0.92}$^{\phantom{\dagger}}$\scriptsize{(\textcolor{dark_green}{$\Uparrow$.10})}& 0.71$^{\phantom{\dagger}}$\scriptsize{(\textcolor{dark_green}{$\Uparrow$.17})}& 0.71 & 0.80 \\ \cmidrule(l){2-17}       
                      & \textbf{\textit{SEEED}}       & \textbf{0.83} & 0.94$^{\phantom{\dagger}}$\scriptsize{(\textcolor{dark_green}{$\Uparrow$.07})} & \textbf{0.75}$^\mathbf{\dagger}$\scriptsize{(\textcolor{dark_green}{$\Uparrow$.27})}& 0.74 & 0.91 & \textbf{0.79} & \textbf{0.94}$^{\phantom{\dagger}}$\scriptsize{(\textcolor{dark_green}{$\Uparrow$.14})}& \textbf{0.68}$^\mathbf{\dagger}$\scriptsize{(\textcolor{dark_green}{$\Uparrow$.23})}& \textbf{0.69} & \textbf{0.87} & \textbf{0.89} & 0.90$^{\phantom{\dagger}}$\scriptsize{(\textcolor{dark_green}{$\Uparrow$.08})}& \textbf{0.87}$^\mathbf{\dagger}$\scriptsize{(\textcolor{dark_green}{$\Uparrow$.33})}& \textbf{0.78} & \textbf{0.84} \\ \midrule
\multirow{4}{*}{75\%} & KNN-Contrastive     & 0.63 & 0.85$^{\phantom{\dagger}}$\scriptsize\phantom{($\Uparrow$.00)}& 0.50$^{\phantom{\dagger}}$\scriptsize\phantom{($\Uparrow$.00)}& 0.49 & 0.82 & 0.44 & 0.85$^{\phantom{\dagger}}$\scriptsize\phantom{($\Uparrow$.00)}& 0.29$^{\phantom{\dagger}}$\scriptsize\phantom{($\Uparrow$.00)}& 0.33 & 0.72 & 0.57 & 0.81$^{\phantom{\dagger}}$\scriptsize\phantom{($\Uparrow$.00)}& 0.43$^{\phantom{\dagger}}$\scriptsize\phantom{($\Uparrow$.00)}& 0.34 & 0.52 \\ 
                      & SynCID              & 0.73 & 0.89$^{\phantom{\dagger}}$\scriptsize{(\textcolor{dark_green}{$\Uparrow$.04})}& 0.62$^{\phantom{\dagger}}$\scriptsize{(\textcolor{dark_green}{$\Uparrow$.12})}& 0.60 & 0.86 & 0.63 & 0.85$^{\phantom{\dagger}}$\scriptsize\phantom{{(\textcolor{dark_green}{$\Uparrow$.04})}}& 0.50$^{\phantom{\dagger}}$\scriptsize{(\textcolor{dark_green}{$\Uparrow$.21})}& 0.47 & 0.78 & 0.66 & 0.78$^{\phantom{\dagger}}$\scriptsize{(\textcolor{dark_red}{$\Downarrow$.03})}& 0.57$^{\phantom{\dagger}}$\scriptsize{(\textcolor{dark_green}{$\Uparrow$.14})}& 0.40 & 0.60 \\ 
                      & LOOP                & 0.79 & 0.92$^{\phantom{\dagger}}$\scriptsize{(\textcolor{dark_green}{$\Uparrow$.07})} & 0.68$^{\phantom{\dagger}}$\scriptsize{(\textcolor{dark_green}{$\Uparrow$.18})}& 0.68 & \textbf{0.90} & 0.64 & 0.87$^{\phantom{\dagger}}$\scriptsize{(\textcolor{dark_green}{$\Uparrow$.02})} & 0.51$^{\phantom{\dagger}}$\scriptsize{(\textcolor{dark_green}{$\Uparrow$.22})} & 0.50 & 0.81 & 0.76 & 0.92$^{\phantom{\dagger}}$\scriptsize{(\textcolor{dark_green}{$\Uparrow$.11})} & 0.64$^{\phantom{\dagger}}$\scriptsize{(\textcolor{dark_green}{$\Uparrow$.21})} & 0.57 & 0.72 \\ \cmidrule(l){2-17}       
                      & \textbf{\textit{SEEED}}       & \textbf{0.87} & \textbf{0.97}$^\mathbf{\dagger}$\scriptsize{(\textcolor{dark_green}{$\Uparrow$.12})} &\textbf{0.78}$^\mathbf{\dagger}$\scriptsize{(\textcolor{dark_green}{$\Uparrow$.28})} & \textbf{0.72} & \textbf{0.90} & \textbf{0.79} & \textbf{0.93}$^{\phantom{\dagger}}$\scriptsize{(\textcolor{dark_green}{$\Uparrow$.08})} &$^{\phantom{\dagger}}$\textbf{0.69}$^\mathbf{\dagger}$\scriptsize{(\textcolor{dark_green}{$\Uparrow$.40})}& \textbf{0.60} & \textbf{0.82} & \textbf{0.86} &$^{\phantom{\dagger}}$\textbf{0.97} \scriptsize{(\textcolor{dark_green}{$\Uparrow$.16})}& \textbf{0.77}$^\mathbf{\dagger}$\scriptsize{(\textcolor{dark_green}{$\Uparrow$.34})}& \textbf{0.71} & \textbf{0.78} \\ \bottomrule
\end{tabular}
}